\PassOptionsToPackage{bookmarksnumbered,unicode}{hyperref}
\documentclass[sigconf, nonacm, pdfa]{acmart}
\usepackage[a-2b]{pdfx}
\usepackage{ragged2e}         
\newcommand\vldbdoi{10.14778/3796195.3796214}
\newcommand\vldbpages{1046 - 1059}
\newcommand\vldbvolume{19}
\newcommand\vldbissue{5}
\newcommand\vldbyear{2026}
\newcommand\vldbauthors{\authors}
\newcommand\vldbtitle{\shorttitle}
\newcommand\vldbavailabilityurl{https://github.com/omnia-postech/OrbitFlow}
\newcommand\vldbpagestyle{empty}

\usepackage{algorithm, algpseudocode}
\usepackage{amsmath}
\usepackage{mathtools}
\usepackage{url}
\usepackage{epstopdf}
\usepackage{tabularx}
\usepackage{enumitem}	
\usepackage{listings}
\usepackage{changepage}
\usepackage{pifont}
\usepackage{multirow}
\usepackage{wrapfig}
\usepackage{color,soul}
\usepackage{array,booktabs}
\usepackage[normalem]{ulem}
\usepackage{xspace}
\usepackage{graphicx}
\usepackage{multicol}
\usepackage{siunitx}
\usepackage{caption}
\usepackage{subcaption}
\usepackage{balance}
\usepackage{epsfig}
\usepackage{tabu}
\usepackage{tikz}
\usepackage{xcolor}
\usepackage{fancyhdr}
\usepackage[T1]{fontenc}

\makeatletter
\@ifpackageloaded{xcolor}{}{\usepackage{xcolor}}
\@ifpackageloaded{enumitem}{}{\usepackage{enumitem}}
\@ifpackageloaded{multicol}{}{\usepackage{multicol}}
\@ifpackageloaded{tcolorbox}{}{\usepackage{tcolorbox}\tcbuselibrary{skins,breakable}}
\makeatother

\newtcolorbox{reviewbox}[1][]{%
  enhanced, breakable,
  colback=gray!10, colframe=gray!60!black,
  boxrule=0.5pt, arc=2pt, outer arc=2pt,
  left=5pt, right=5pt, top=3pt, bottom=3pt,
  fonttitle=\bfseries,
  title=Reviewer Comment, #1
}

\makeatletter
\renewcommand\paragraph{\@startsection{paragraph}{4}{\z@}%
	{-.3ex \@plus -.3ex \@minus -.0ex}%
	{-1.ex}%
	{\normalfont\bfseries}}
\makeatother

\newcommand{\Paragraph}[1]{\vskip 3pt\noindent\textbf{#1 }}

\newenvironment{myitemize}%
  {\begin{itemize}
	[leftmargin=0cm,
		itemindent=.3cm,
		labelwidth=\itemindent,
		labelsep=0pt,
		parsep=3pt,
		topsep=2pt,
		itemsep=1pt,
		align=left]
  }%
  {\end{itemize}}

\newenvironment{myenumerate}%
  {\begin{enumerate}
	[leftmargin=0cm,itemindent=.5cm,labelwidth=\itemindent,
		labelsep=0pt,
		parsep=1pt,
		topsep=1pt,
		itemsep=3pt,
		align=left]
  }%
  {\end{enumerate}}

\def\sectionautorefname{$\S$}
\def\subsectionautorefname{$\S$}
\def\subsubsectionautorefname{$\S$}

\makeatletter
\newcommand{\thickhline}{%
    \noalign {\ifnum 0=`}\fi \hrule height 1pt
    \futurelet \reserved@a \@xhline
}
\newcolumntype{"}{@{\hskip\tabcolsep\vrule width 1pt\hskip\tabcolsep}}
\makeatother

\newcommand{\eat}[1]{}

\newcommand{\ours}{{\textsc{OrbitFlow}}}

\renewcommand{\sectionautorefname}{§\kern-2.5pt}
\renewcommand{\subsectionautorefname}{§\kern-2.5pt}
\renewcommand{\subsubsectionautorefname}{§\kern-2.5pt}

\newcommand{\revmajor}[1]{{\color{black}{#1}}}
\newcommand{\revminor}[1]{{\color{black}{#1}}}
\newcommand{\revrewrite}[1]{{\color{black}{#1}}}

\begin{document}

\date{}


\pagenumbering{arabic}   
\title{OrbitFlow: SLO-Aware Long-Context LLM Serving with Fine-Grained KV Cache Reconfiguration}

\author{Xinyue Ma}
\authornote{These authors contributed equally to this work.}
\affiliation{
  \institution{POSTECH}
}
\email{xinyuema@postech.ac.kr}

\author{Heelim Hong}
\authornotemark[1]
\affiliation{
  \institution{UNIST}
}
\email{heelim@unist.ac.kr}

\author{Taegeon Um}
\affiliation{
  \institution{Samsung Research}
}
\email{taegeon.um@samsung.com}

\author{Jongseop Lee}
\affiliation{
  \institution{POSTECH}
}
\email{jslee202403@postech.ac.kr}

\author{Seoyeong Choy}
\affiliation{
  \institution{POSTECH}
}
\email{sychoy0704@postech.ac.kr}

\author{Woo-Yeon Lee}
\affiliation{
  \institution{Samsung Research}
}
\email{wooyeon0.lee@samsung.com}

\author{Myeongjae Jeon}
\affiliation{
  \institution{POSTECH}
}
\email{mj.jeon@postech.ac.kr}

\begin{abstract}

Serving long-context LLMs is challenging because request lengths and batch composition vary during token generation, causing the memory footprint to fluctuate significantly at runtime. 
Offloading KV caches to host memory limits effective GPU memory usage, but existing static and predetermined offloading strategies cannot adapt to the rapidly shifting memory demands of long-context serving. This often leads to excessive CPU-to-GPU KV transfers that translate into latency spikes and frequent SLO violations.

To address these challenges, we introduce \ours{}, a fine-grained and adaptive KV cache management system that meets latency SLOs in long-context LLM serving. \ours{} employs a lightweight ILP solver to decide which layers' KV caches to retain on the GPU for each request, within memory capacity constraints. It continuously refines KV placements based on runtime feedback when the active plan becomes suboptimal during token generation. Under heavy load, \ours{} invokes a fallback mechanism to temporarily defer in-flight requests with large memory footprints, preserving overall SLO attainment. Our experiments demonstrate that \ours{} improves SLO attainment for TPOT and TBT by 62\% and 66\%, respectively, while reducing the 95th percentile (i.e., P95) latency by 38\% and achieving up to $3.3\times$ higher throughput compared to existing offloading methods.


\end{abstract}

\maketitle

\pagestyle{\vldbpagestyle}
\begingroup\small\noindent\raggedright\textbf{PVLDB Reference Format:}\\
\vldbauthors. \vldbtitle. PVLDB, \vldbvolume(\vldbissue): \vldbpages, \vldbyear.\\
\href{https://doi.org/\vldbdoi}{doi:\vldbdoi}
\endgroup
\begingroup
\renewcommand\thefootnote{}\footnote{\noindent
This work is licensed under the Creative Commons BY-NC-ND 4.0 International License. Visit \url{https://creativecommons.org/licenses/by-nc-nd/4.0/} to view a copy of this license. For any use beyond those covered by this license, obtain permission by emailing \href{mailto:info@vldb.org}{info@vldb.org}. Copyright is held by the owner/author(s). Publication rights licensed to the VLDB Endowment. \\
\raggedright Proceedings of the VLDB Endowment, Vol. \vldbvolume, No. \vldbissue\ %
ISSN 2150-8097. \\
\href{https://doi.org/\vldbdoi}{doi:\vldbdoi} \\
}\addtocounter{footnote}{-1}\endgroup

\ifdefempty{\vldbavailabilityurl}{}{
\begingroup\small\noindent\raggedright\textbf{PVLDB Artifact Availability:}\\
The source code, data, and/or other artifacts have been made available at \url{\vldbavailabilityurl}.
\endgroup
}

\section{Introduction}
\label{sec:Introduction}
Large language models (LLMs) have gained significant prominence across
various natural language processing tasks, including text generation~\cite{deepspeed, gpt3, llama3, palm, acm2023con, questiongen2024acl},
language translation~\cite{multilingual, mt5, llamax}, summarization~\cite{bart, pegasus}, and question
answering~\cite{rag, chain_thought, open_gen}. Their capabilities have greatly improved over the past few
years, largely empowered by dramatic increases in \emph{context window}
size---i.e., the maximum number of tokens in words
a model can
process at once. For instance, GPT-2~\cite{gpt2} supported a context
window of 1K tokens, \revrewrite{while its successor GPT-3~\cite{gpt3} doubled the window size}
to 2K
tokens. More recent models such as GPT-4~\cite{gpt4} and Claude~\cite{claude} have
\revrewrite{pushed the limits}
further, handling up to 128K or even
1M tokens.

To provide the best user experience, LLM serving systems must respond to requests within stringent latency requirements defined by
service-level objectives (SLOs)~\cite{distserve, splitwise, sarathiserve}.
Upon receiving an input sequence of tokens,
\revrewrite{these systems}
generate output tokens autoregressively, one token at a time, through
consecutive \emph{decode steps}. 
Many LLM services, such as chat agents, creative writing tools, and real-time 
translators, stream tokens as they are generated \revrewrite{so that} users can read responses incrementally. 
To keep decode latency perceptually invisible, these services typically set \revrewrite{per-token} latency SLOs (e.g.,
200--300~ms for chatbots) to match natural reading speeds~\cite{readingspeed, distserve}. Frequent
violations of these latency SLOs can impair user experience and reduce
revenue~\cite{slo_quality}, regardless of response quality. \revrewrite{As context windows expand and workloads become more decode-dominant, optimizing for per-token latency SLOs becomes ever more critical.}

\revrewrite{However, serving long-context LLMs} within strict time bounds is challenging because GPU memory demands
vary widely during runtime. Modern
serving systems \revrewrite{accelerate token generation by} performing decode steps based on cached states of all preceding
input-output tokens, termed \emph{KV cache}. 
Rapid decoding thus requires storing all KV caches
in GPU memory, which was manageable \revrewrite{with} small context windows.
\revrewrite{Such caching has now become nontrivial for two reasons:} (1) although initially small, the KV cache grows \revrewrite{steadily during lengthy} decoding, creating a high likelihood of exhausting GPU memory (i.e., \emph{token dimension}); and (2) LLM serving systems often dynamically batch multiple requests to
process them concurrently on the GPU~\cite{vllm, orca}, further straining the
available GPU memory (i.e., \emph{batch dimension}).
There are thriving efforts to reduce KV cache size itself \revrewrite{during the decode step. For instance,} state restoration~\cite{hcache} \revrewrite{recomputes parts of the KV cache
at runtime}, and token pruning~\cite{lazyllm, quest,
streamllm, h2o, infinigen} \revrewrite{discards presumably less
critical KV entries}. While effective,
these approaches
carry inherent limitations, as they
either apply only to
specific models with multi-head attention (MHA)~\cite{hcache} or risk
degrading response quality.

This paper tackles GPU memory 
\revrewrite{challenges} in long-context serving through \emph{KV offloading}~\cite{layerkv}, which moves parts of the KV cache to host memory and \revrewrite{overlaps their retrieval with ongoing GPU computation.} 
\revrewrite{KV offloading does not compromise accuracy and, when scheduled concurrently, can achieve effective computation-communication overlap. 
The key insight behind this is that each layer uses only its own cached KV states and does not access caches from other layers. 
However, current common practices are largely SLO-unaware:} they statically and monolithically offload and prefetch states for all batched requests layer by layer \revrewrite{to save memory~\cite{flexgen, infinigen}. Even with larger offload distances, i.e., evicting caches at regular but wider layer intervals within the GPU memory budget, static strategies fail to conceal escalating data-transfer overheads as the cache expands} along both token and
batch dimensions.  
This overhead markedly delays GPU
computations and subsequent token decode steps for every request in a batch,
making it difficult to consistently meet latency SLOs across requests.

Our system, \ours{}, realizes effective KV offloading through \emph{adaptive},
\emph{fine-grained} offload plans. At its core, \ours{} employs a lightweight
ILP-based solver that dynamically optimizes GPU memory usage across batched
requests to minimize SLO violations.
The solver iteratively refines optimal offload plans using real-time feedback
on varying state-transfer and GPU computation costs per layer, allowing \ours{}
to quickly adjust decisions under current GPU memory and compute resource
constraints.



For fine-grained KV offloading, \ours{} treats each request in a dynamic
batch individually based on its current stage in the decoding phase. A key
challenge is balancing GPU memory allocation among requests with distinct
memory demands \revrewrite{given limited GPU capacity.} 
For instance, when serving a single request, \ours{} configures the largest
possible offload distance to fully utilize available GPU memory and achieve
maximum token generation speed. When another request arrives later, the two
requests form a batch \revrewrite{at different stages in decoding, with non-identical} KV cache sizes.
Within such batches, the memory demands of requests often conflict:
a newly arrived request requires memory for rapid token generation, whereas
an older request with substantial decoding progress needs to retain a large
portion of its KV cache on GPU to effectively overlap computation and memory
prefetching. To reconcile this, our solver holistically examines
various offload-reload plans within the GPU memory budget at the
\emph{request level} and selects the one that best prevents latency
violations for each request.

It is worth noting that \ours{} may fail to meet latency SLOs when the server
is severely oversubscribed.
Nonetheless, it offers two simple yet effective fallback
mechanisms---\emph{Token-Deposit} and \emph{Pause-Resume}---to mitigate
latency impacts during transient oversubscription. When a request
produces output tokens while aggressively using GPU memory, \ours{} does
not always emit these tokens immediately. Instead, it emits
tokens at a rate compliant with latency SLOs. Specifically, if LLM serving
must adhere to a constant time-between-tokens (TBT)~\cite{sarathiserve,
splitwise}, \ours{} first deposits generated tokens into a buffer and
then releases them at a steady rate matching the predefined TBT to smooth the output rate. This mechanism does not stand alone; our solver uses it to
\revrewrite{temporarily pause} suitable requests with high memory footprints and many buffered tokens, freeing GPU memory to better accommodate other
requests in the batch. 
Since \ours{} continues supplying
users with deposited tokens \revrewrite{during pauses}, it maintains the perception of continuous request processing. These mechanisms are particularly beneficial in batches where some requests complete quickly, as resources freed by such requests can then be reclaimed to resume paused requests without disrupting the user experience.

We build \ours{} atop vLLM v0.6.6~\cite{vllm} and evaluate it against baseline offloading systems on ShareGPT-derived synthetic traces across different arrival rates. 
Our evaluation shows that \ours{} improves time-per-output-token (TPOT) and TBT SLO attainment by 62\% and 66\%, respectively, 
while reducing P95 latency by 38\% \revrewrite{and achieving up to 3.3$\times$ higher throughput, over existing methods.} 
Additionally, \ours{} consistently provides performance improvements across different context lengths and batch sizes, and under distributed serving scenarios.


\section{KV Offloading: A Unified View}
\label{sec:background}
\revrewrite{Service-level objectives (SLOs) typically target millisecond-level latency.
Yet, the KV cache that grows linearly in batched long-context workloads and
exhausts GPU memory makes these SLOs hard to meet
with existing static KV offloading 
methods~\cite{layerkv, flexgen}.
In this section, we detail these problems and explain why 
dynamic offloading is particularly necessary for long-context LLM inference.}

\subsection{Background}
\label{sec:2_1}

\begin{figure}[t]
	\centering
	\includegraphics[width=0.47\textwidth]{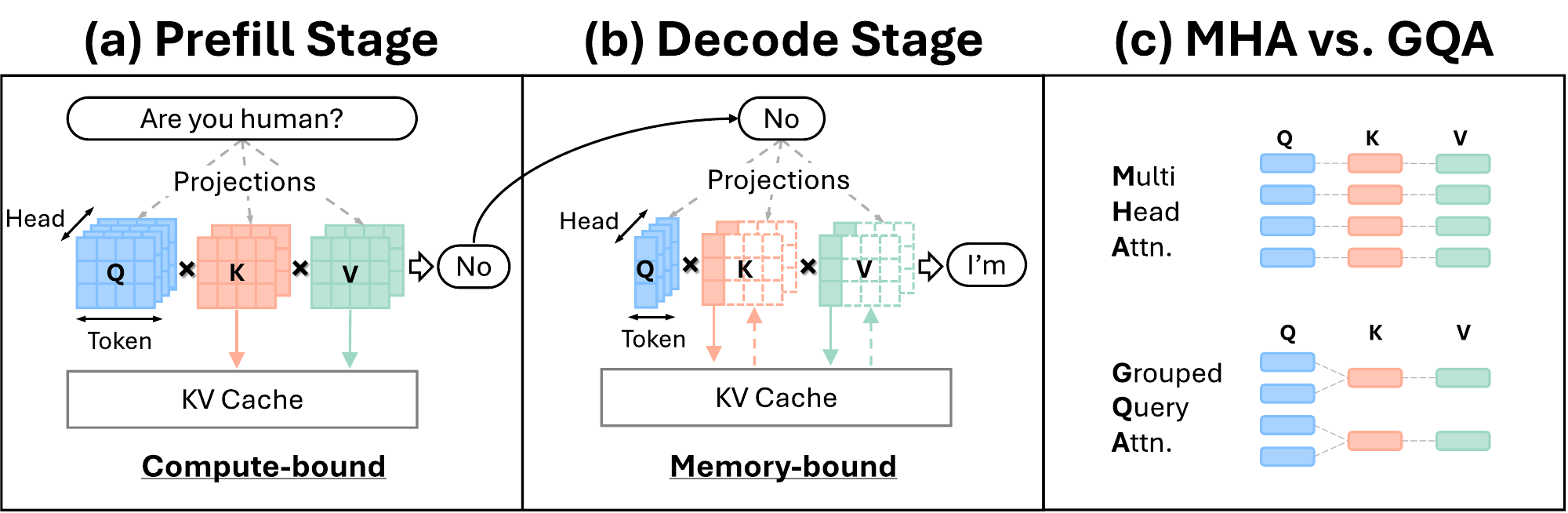}
	\vspace{-0.3em}
	\caption{\revmajor{(a) Prefill stage computes all input tokens in parallel. (b) Decode stage loads KV from the cache and computes only one token at a time. (c) MHA maps distinct KV per query head, while GQA shares KV across head groups to reduce the KV size.} }
    \Description{}
	\label{fig:prefill_decode} 
	\vspace{-0.5em}
\end{figure}

Many contemporary LLM workloads involve decode-dominant, highly interactive tasks
like chat agents, code assistants, and creative writing tools. Such tasks often require context lengths exceeding 32K tokens, sometimes up to millions, as seen in recent flagship models~\cite{gpt4,longt5,llama3,claude,gemini,longnet, yi, qwen, internlm, mistral}. The KV size grows with the input-output sequence length, so longer contexts translate directly into higher memory demand.

\paragraph{Inference Stages.} In modern LLMs built from stacked Transformer layers, \revrewrite{inference comprises two
distinct stages, as shown in~\autoref{fig:prefill_decode}: \emph{prefill} (compute-bound),} where input tokens are
processed in parallel to generate initial context information, \revrewrite{and
\emph{decode} (memory-bound),} where tokens are generated autoregressively.

\begin{figure}[t]
	\centering
	\includegraphics[width=0.47\textwidth]{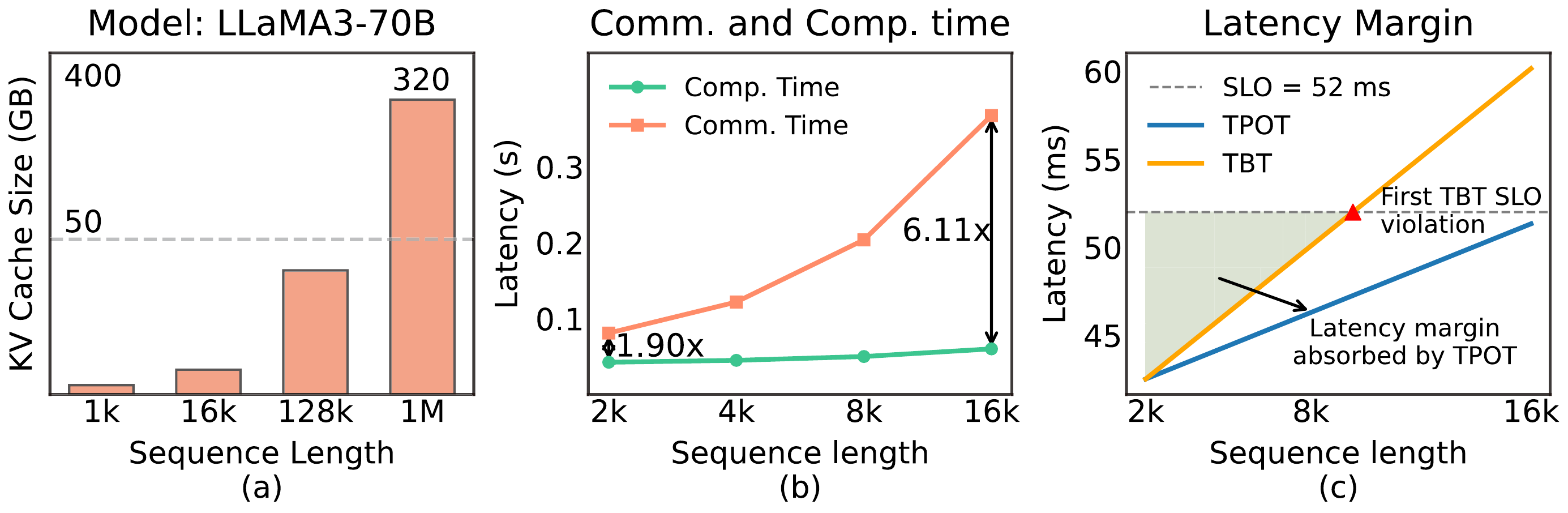}
	\caption{(a) The KV cache size of various sequence lengths with the LLaMA3-70B model. 
    \revrewrite{(b) A layer's computation time and communication time to transfer its KV from CPU to GPU over sequence lengths with the LLaMA3-8B model.}
    (c) With a fixed token-level SLO, tokens generated in early steps become the latency margin.
    }
    \Description{}
	\label{fig:kv_cache_com_comp} 
	\vspace{-0.2em}
\end{figure}

\revmajor{At the core of both stages is the attention mechanism. In this mechanism, inputs are projected to query (Q), key (K), and value (V) tensors, which are split into multiple heads for parallel processing. As shown in \autoref{fig:prefill_decode}c, unlike multi-head attention (MHA) adopted by initial GPT models (e.g., GPT-2), recent models such as LLaMA3 adopt grouped-query attention (GQA), where the KV tensors are split into fewer heads than the Q tensor to reduce the KV cache size.}
\revrewrite{Even with GQA, the KV size remains massive 
at sequence lengths from 128K to 1M tokens,
as shown in~\autoref{fig:kv_cache_com_comp}a.
}

\paragraph{SLO Metrics.}
In LLM serving, time-to-first-token (TTFT) and
time-per-output-token (TPOT) are common SLO metrics because they provide straightforward
insights into user-perceived responsiveness and average throughput. On their own, however,
these metrics may not fully capture critical aspects of
latency \revrewrite{in} extended context scenarios. While TPOT accurately
measures average latency, it does not reveal whether token generation is
consistently steady, since slow tokens can be offset by faster ones. \revrewrite{As a result,} recent research highlights token-level latency metrics, such as
time-between-tokens (TBT), alongside TPOT. 
\revrewrite{For the same latency target} (e.g., 50 ms per token),
under a TPOT SLO, a request violates the SLO when its average per-token latency exceeds the target, whereas under a TBT SLO, \revrewrite{each token exceeding that latency constitutes a violation.}
TBT thus explicitly pinpoints individual token delays that affect user experience in interactive
workloads~\cite{sarathiserve,splitwise,jha2024learned}. By adopting both
TPOT and TBT, we gain a comprehensive view of system responsiveness, ensuring that both overall throughput and fine-grained latency variations are captured and evaluated.

\begin{figure*}
    \centering
    \includegraphics[width=0.98\linewidth]{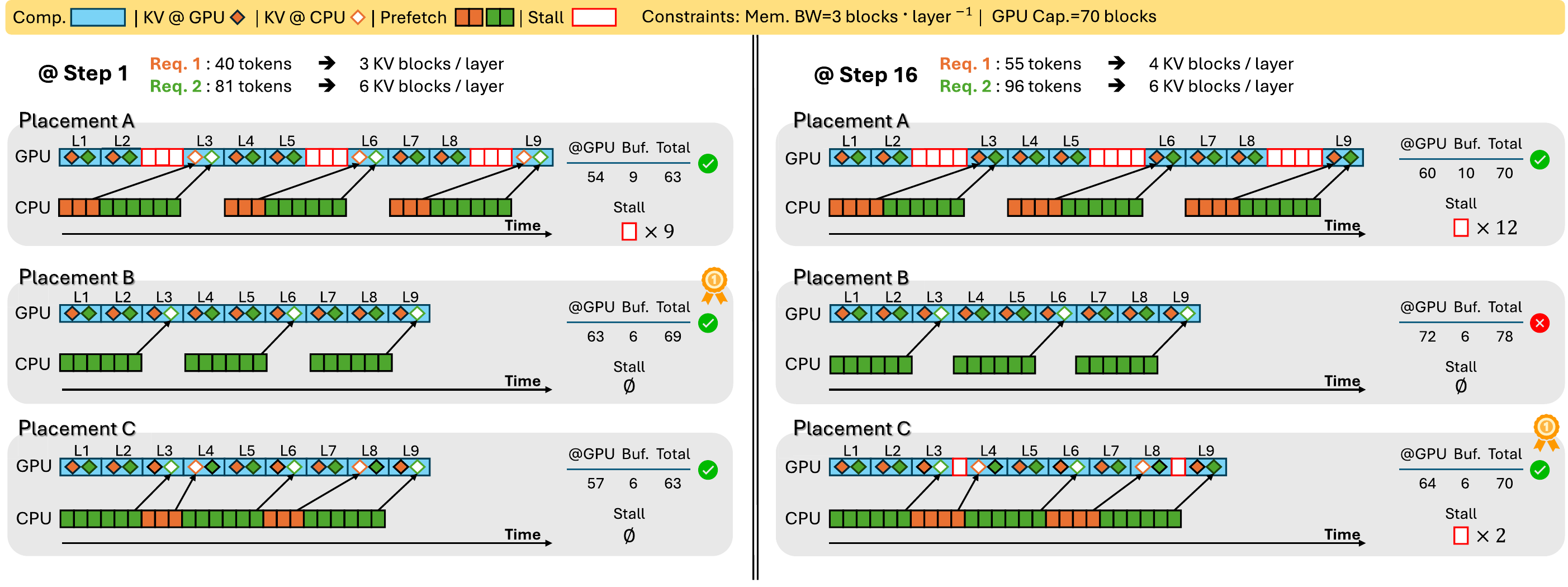}
    \caption{An illustrative example showing why a static, uniform offload policy is
insufficient. The timelines track one decode step early
    (Step~1, left) and one much later (Step~16, right) for three
    placements.  Each timeline is annotated with the resident KV blocks
    on GPU, the size of the per-layer prefetch buffer, and the resulting
    stalls. 
    }
    \Description{}
    \label{fig:toy_example}
\end{figure*}
\label{sec:deposit}

\paragraph{Continuous Batching.} \revrewrite{Production LLM servers process
multiple concurrent requests by grouping them into batches. Continuous
batching~\cite{orca, vllm, sarathiserve, distserve, vtc} is a widely adopted
strategy because it sustains high throughput without provisioning additional
serving instances. Under continuous batching, when any request in a batch
completes, a new request is admitted and its prefill begins, while the
ongoing decoding is temporarily paused. Unless otherwise specified, we assume
continuous batching for all subsequent analyses.}


\subsection{Drawbacks of Previous Offloading Methods}
\label{sec:prev_methods}
Previous offloading methods treat all requests within a batch uniformly~\cite{deepspeed-inference,flexgen,selectn}. They operate at the layer granularity: if a layer is offloaded, the KV entries for that layer are moved to the CPU for all requests; \revrewrite{otherwise, all these KV entries stay} resident on GPU. On top of this, the layers to offload are selected in a distance-driven manner, \revrewrite{meaning they are evenly} distributed across the model depth. For instance, if two out of six layers are offloaded, the offload pattern would look like [1 1 0 1 1 0], where 1 denotes a layer kept on GPU and 0 denotes one offloaded to CPU.

\revrewrite{In Transformer-based LLM inference}, computation in a given layer can start only after its entire KV cache is present on the GPU. \revrewrite{Therefore, each layer's KV cache across batched requests share a single operational timeline.} The computation time per layer is consistent, with the same amount of KV \revrewrite{data to transfer if a layer is offloaded.} This makes offloading every ${\mathrm{N}}$-th layer of a model equivalent to repeating groups of ${\mathrm{N}}$ layers \revrewrite{(e.g., two groups of [1 1 0] in the example above)}. Within each group, the computation, communication, and resulting stalls are identical, giving the runtime a steady and predictable pattern to exploit.

The distance ${\mathrm{N}}$ is usually chosen based on the worst-expected
workload, so that KV caches at maximum sequence length still fit in GPU memory
and \revrewrite{offloaded layers can overlap their transfers with computation as much as
possible.} Some prior systems~\cite{flexgen, selectn} \revrewrite{profile offloading
distances over a grid of batch sizes and sequence lengths to select the best
value for each workload. Fixing this distance ahead of time}
avoids runtime overhead but can be suboptimal if actual workloads deviate
from the worst-case assumptions. \revrewrite{This reduces} compute-communication overlap
and increases stall risk. We examine this effect in two representative cases:
token-dimension drift and batch-dimension mismatch.

\Paragraph{Token-Dimension Drift.}
\autoref{fig:kv_cache_com_comp}b plots
computation and data transfer times against sequence length for LLaMA3-8B on
a single NVIDIA A6000 GPU. As decoding proceeds \revrewrite{and sequence length grows,
both} costs grow linearly. However, the \revrewrite{transfer time rises} faster, making the
optimal offloading strategy dependent on \revrewrite{their relative} ratio. For instance,
at a sequence length of 2K, data transfer takes about 1.9$\times$ the time
needed to compute one layer. In this case, offloading every third layer hides
the transfer time under computation, yielding perfect overlap. As
decoding continues and the sequence length reaches 16K, the ratio becomes
6.1$\times$, and the offloading strategy that worked at 2K tokens no longer
fully hides the transfer time and must be adjusted.

Notably, reducing decode latency always improves the TPOT SLO attainment, but does not always improve TBT SLO attainment. For example, if we 
apply an offloading strategy optimal for 
perfect overlap and the sequence is initially short,
early tokens 
may be emitted faster than necessary. However, each subsequent token's TBT SLO resets relative to the prior one, so early savings cannot be carried forward in improving TBT attainment. 
To illustrate, the green area in~\autoref{fig:kv_cache_com_comp}c 
represents the \emph{latency margin} created while the TBT stays below the 52 ms SLO.
\revrewrite{As the sequence length increases,} the TBT 
\revrewrite{grows more rapidly than the TPOT}. Around 9K tokens, the margin is gone even though TPOT stays low, showing that early speed gains improve TPOT but not TBT.
In this work, we track \emph{latency margin} as a system-wide resource, harvesting it early and spending it to mask later stalls. 

\Paragraph{Batch-Dimension Mismatch.}
Real-world workloads vary substantially in request lengths, arrival times, and output sizes. As a result, requests in the same batch may be at different decoding stages with different KV cache sizes. Compute time grows with the total number of tokens in the batch, and data transfer time grows with the number of KV blocks that are offloaded. Under a uniform distance-driven offload policy, the KV cache of one layer is either entirely on GPU or CPU, so both costs scale with the total KV size rather than the individual KV cache sizes of each request. Short requests are therefore delayed, while the KV of a long request pushes up both compute and data transfer time for the batch.

This impact also depends on the batching strategy. Consider a batch with one long request processed alongside several short requests, with additional short requests waiting in the queue.
Continuous batching allows the admission of one queued request whenever a short request in the batch finishes. The long request, however, continues to dominate compute and data transfer and slows down token generation. 
\revrewrite{This is a poor trade-off: a single request with a large KV footprint can put all other short requests at risk of SLO violations, thereby degrading overall system performance.}

\subsection{An Illustrative Example}
\label{sec:illustrative_example}

To ground the two limitations pinpointed in the previous section, we
walk through the minimal scenario shown in~\autoref{fig:toy_example}. \revmajor{We assume the PagedAttention KV layout~\cite{vllm}, where each layer's KV cache is allocated in fixed-size blocks of 16 tokens.} Two concurrent requests are served on a 9-layer LLM with Request 2 around twice as long, highlighting the \emph{batch-dimension mismatch}. The left half of the figure captures this initial state (Step 1), while the right half shows the status after 15 decode steps (Step 16). By this point, Request 1's KV cache has grown from three to four blocks per layer because its additional tokens demand a new 16-token \revmajor{block---each decode step adds one token, but allocation increases only when a 16-token block boundary is crossed}. Request 2, however, still fits within its existing blocks, as shown in the figure's upper panel. This growth illustrates \emph{token-dimension drift}. 
In this example, we assume the GPU can store at most 70 KV blocks, and the PCIe bandwidth allows transferring at
most 3 KV blocks within the computation time of one layer. When the combined KV cache size
of all requests exceeds this budget, some blocks must stay in CPU memory
and be prefetched into a small staging buffer on the GPU to process the corresponding layer.

\revrewrite{The scenario in~\autoref{fig:toy_example} compares three KV placement plans.}
\textbf{Placement A} serves as an example of the \emph{uniform} offload method in which both requests offload the evenly spaced layers L3, L6 and L9. \textbf{Placement B} attempts to mitigate the batch-dimension mismatch by keeping all KVs of the shorter Request 1 on the GPU and offloading 3 layers for the longer Request 2. \textbf{Placement C} offloads 2 layers of Request 1 and 3 layers of Request 2, trading a little more CPU-to-GPU traffic for extra headroom in GPU memory.

\paragraph{At Decoding Step 1.} Placement A must move
\textbf{9}~blocks just in time for every offloaded layer. Since the bandwidth allows transferring 3 blocks per layer time, those transfers spill past the preceding layer's
compute and create 9 units of stalls. Placement A is already the \emph{best} uniform choice: offloading one more layer leaves even less computation time to overlap with the data transfer, resulting in longer stalls; offloading one less layer requires keeping too many blocks on GPU, violating the 70-block memory limit. Both Placements B and C apply request-wise placement and achieve perfect overlap without stalls, but Placement B transfers fewer blocks and keeps higher GPU memory utilization. 

\vspace{0.5\baselineskip} 
\noindent\fbox{%
  \parbox{0.96\linewidth}{%
    \textbf{Observation 1)} 
    A uniform offload policy that ignores the batch-dimension mismatch can miss the optimal placement and leave avoidable stalls. Allowing \emph{fine-grained} per-request placements can better overlap computation with data transfer.%
  }%
}

\paragraph{At Decoding Step 16.} After 15 more decode steps, Request 2 still needs 6 blocks per layer, while Request 1 has grown to 4 blocks per layer.  These additional blocks push Placement B's memory usage to 78 blocks---\emph{beyond} the 70-block GPU memory capacity---making it infeasible. Placement A remains feasible but must now prefetch 10 blocks per layer, causing 12 stalls in total. Placement C, which had sacrificed a bit of GPU residency earlier, still fits within capacity
and incurs only
2 stalls, becoming the new optimal strategy. 

\vspace{0.5\baselineskip} 
\noindent
\fbox{%
  \parbox{0.96\linewidth}{%
    \textbf{Observation 2)} A placement that was optimal early in decoding can later become suboptimal---or even infeasible---as KV caches grow. A single, static placement therefore cannot serve all decode steps; \emph{dynamic reconfiguration} is required.
  }%
}
\section{System Design}
\label{sec:design}

\begin{figure}[t]
	\centering
	\includegraphics[width=0.45\textwidth]{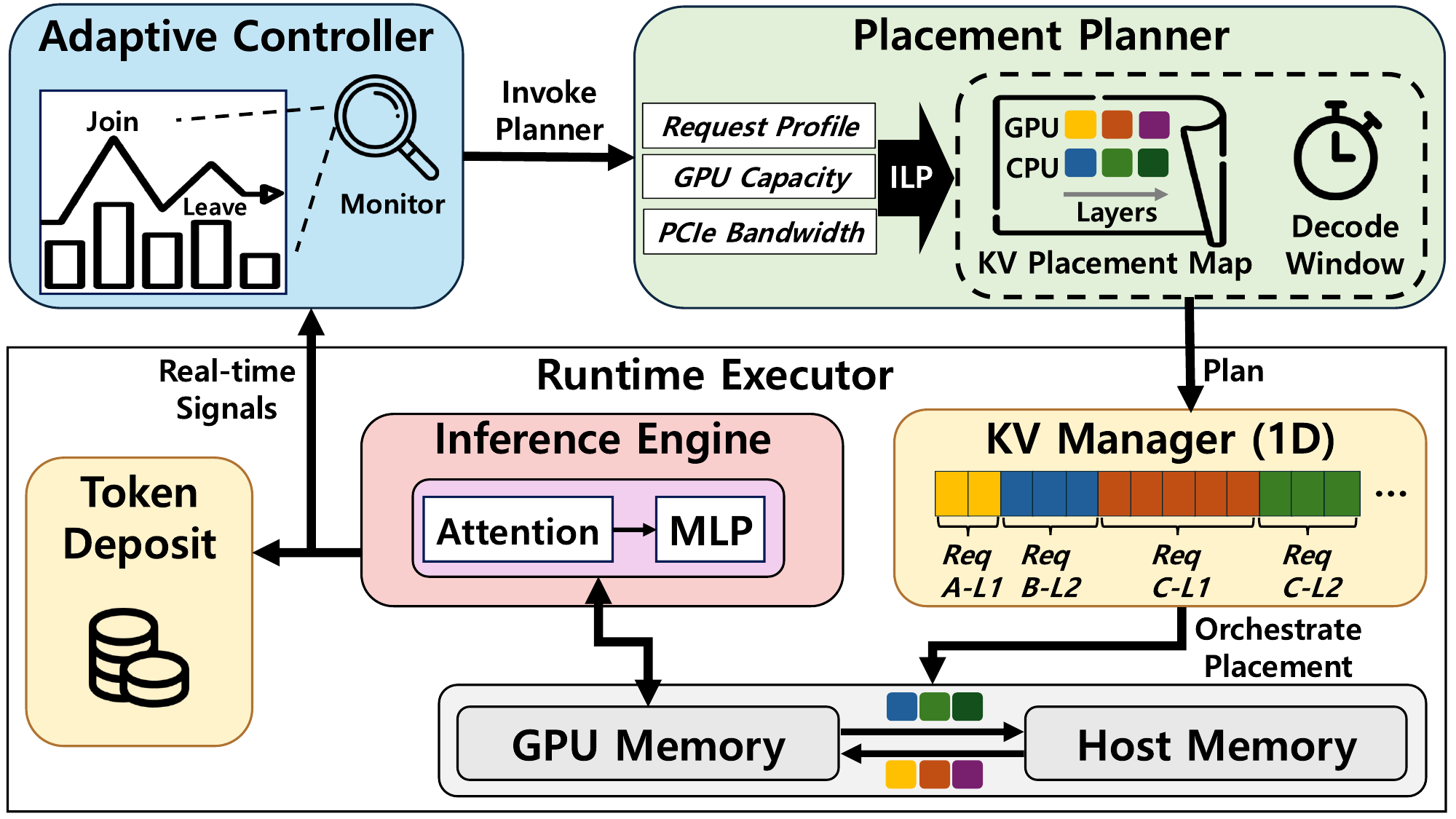}
	\vspace{-0.3em}
	\caption{The overall architecture of \ours. }
    \Description{}
	\label{fig:arch} 
	\vspace{-0.5em}
\end{figure}

We now introduce \ours{}, an implementation of fine-grained, adaptive KV cache placement that optimizes KV offloading for long-context LLM serving under limited GPU memory.

\revmajor{\subsection{Design Overview}}

\revmajor{The design of \ours{} is guided by the following principles:}

\revmajor{
\begin{myitemize}
    \item \textbf{Fine-grained Placements.} To enable fine-grained offloading, \ours{} treats each request-layer combination as a placement choice on either GPU or host memory and accounts for how transfer schedules interact with per-layer compute costs. Because decode steps last only tens of milliseconds, the solver must stay lightweight by exploring the search space efficiently without violating SLOs.
    \item \textbf{Dynamic Reconfiguration.} The system must decide when to reconfigure to balance stability and adaptivity. Reconfiguring after every step closely tracks workload drifts, but frequent data movement and plan switching can offset gains. In contrast, infrequent reconfiguration risks performance degradation and SLO violations.
    \item \textbf{System-level SLO Management.} To meet system-level SLOs, it must accumulate latency margins to mask later stalls and, in cases of memory pressure, pause a long-running, KV-dominant request to guard overall SLOs. It must navigate this global trade-off without inducing starvation, which would affect tail latency.
\end{myitemize}
}

\revmajor{Guided by these principles,} \ours{} 
{operates}
three key components that form a closed-loop control system, shown in \autoref{fig:arch}. 

\textbf{Runtime Executor} (\autoref{sec:executor}) performs each decoding step. It executes computations, \revrewrite{installs a KV placement}, initiates asynchronous KV transfers based on it, and records compute time, transfer time, and GPU memory usage.
To maintain steady user-visible latency in the presence of latency margins, 
\revmajor{tokens generated within the SLO are buffered in the 
\emph{Token-Deposit} and released at the SLO rate.}
The Executor forwards recorded metrics to Adaptive Controller.

\textbf{Adaptive Controller} (\autoref{sec:controller}) evaluates whether the placement remains valid. \revrewrite{It monitors any batch-dimension mismatch that} alters the KV cache size and the ratio between compute and communication, as well as the current placement's expiry time provided by the Placement Planner, 
which indicates when
token-dimension drift is expected to invalidate the current placement. If any of these conditions is triggered, it requests a new placement from the Planner. When 
\revrewrite{the Planner's best placement}
would still violate too many SLOs, the Controller falls back to \emph{Pause-Resume}. 

\textbf{Placement Planner} (\autoref{sec:planner}) \revrewrite{finds the best KV placement given the current request lengths, available GPU block capacity, and the present PCIe bandwidth. It balances SLO awareness, solver overhead, and adaptive call frequency.}
The Planner solves an ILP to choose per-request offload distances that minimize decode latency while respecting memory and SLO constraints. To keep the search space tractable, it applies principled pruning and prevents the exponential explosion that a completely free layout would cause. Solver latency is hidden by running the optimization in parallel with the computation, so the cost does not appear on the critical path. Alongside the new layer-wise KV placement, the Planner also estimates an expiry time---i.e., the specific decoding step at which the growing KV size changes the communication‑to‑computation ratio (\autoref{fig:kv_cache_com_comp}b) or exceeds available GPU memory. This plan is then forwarded to the Executor to install.

\subsection{Runtime Executor}
\label{sec:executor}
Runtime Executor integrates an SLO-aware {Token-Deposit} that decouples token generation from its delivery and a \emph{KV manager} that enforces the KV placements pulled from the \emph{Planner}.

\paragraph{SLO-aware Token-Deposit.}
\label{sec:token_deposit}
Rather than delivering each token as soon as it is generated, \ours{} decouples token generation from its delivery. Doing so allows us to convert any early latency margins into a reserve that can be spent to hide SLO violations. Upon generation, the token is simply appended to {Token-Deposit}. 
While the deposit is non-empty, the Executor delivers exactly one token per SLO interval. An SLO violation is visible only when the deposit is empty. In this case, the next token is delivered as soon as it is generated. If a request is paused or preempted, the deposit continues to drain, hiding the SLO violations until it empties. Once a request finishes, no further violations can occur, and any remaining tokens are delivered in a single burst.

\begin{figure}
    \centering
    \includegraphics[width=0.97\linewidth]{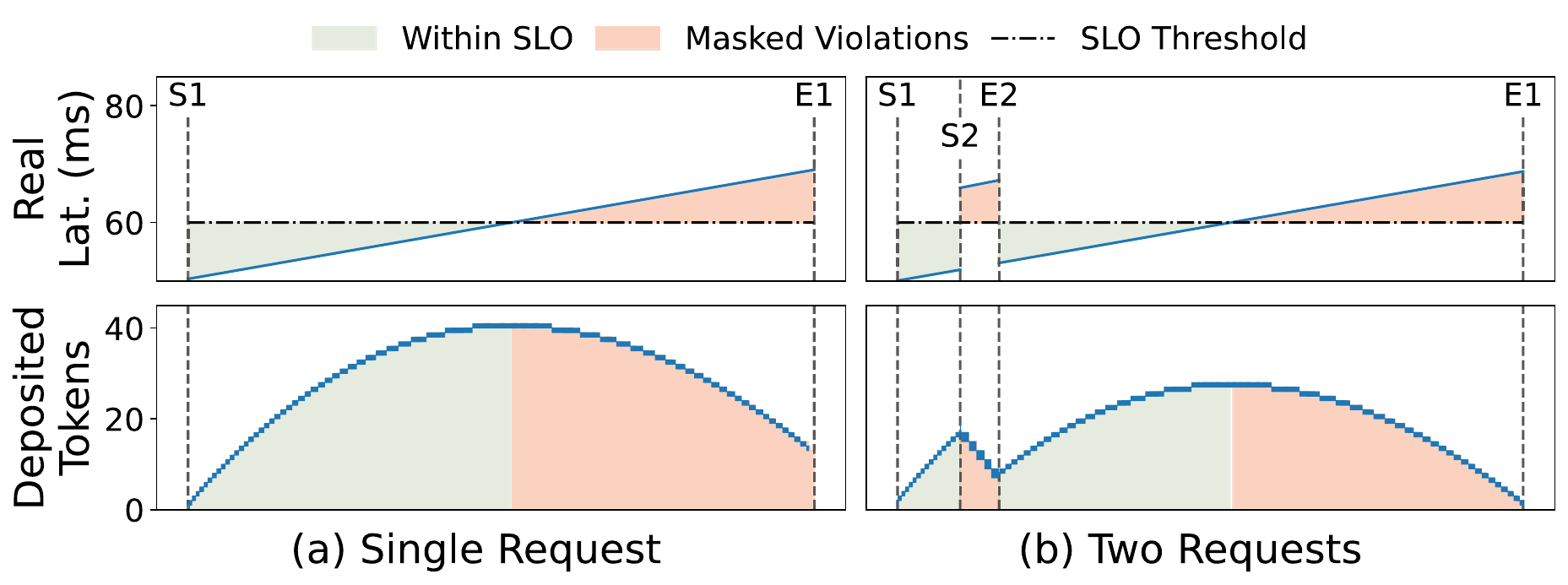}
    \caption{ (a) Deposited tokens mask all the SLO violations for a single request. (b) Deposited tokens mask the transient latency spikes caused by a short-lived request with a long prompt. \revmajor{S1, E1, S2, and E2 mark the start and end of Request 1 and Request 2.}}
    \Description{}
    \label{fig:deposit}
\end{figure}

\autoref{fig:deposit}a illustrates how Token-Deposit helps hide SLO violations for a single request. 
During the early decode iterations, the KV cache is still small, so the generation latency margin remains well below the SLO, and the deposit balance grows steadily. As soon as the generation latency exceeds the SLO, the deposit is drawn down. Although the backend incurs many more SLO violations, the deposit masks a significant portion of them from the user, with no additional computation or memory traffic. The same buffering logic smooths over transient latency spikes, as shown in \autoref{fig:deposit}b. The arrival of a long-prompt, short-output request (\revmajor{i.e., Request 2}) pushes the decode latency over the SLO temporarily, but this spike is mostly absorbed by deposited tokens until it empties. 

\revminor{
The Token-Deposit buffer is a CPU-resident FIFO that stores a 4-byte integer for each generated token. 
Its size is dynamic, but the footprint is negligible, typically only hundreds of KBs. Enqueue and dequeue operations are fast regardless of the buffer size.}

\paragraph{KV Manager.}
\label{sec:kv_manager}
The KV manager \emph{enforces} the placement produced by the solver and keeps it consistent throughout decoding. 
It maintains up-to-date GPU-CPU block tables that record each request's KV cache address and track the next write location for every decode iteration.
When a new placement arrives, the manager computes the delta with respect to the current layout and performs the minimum data movement needed to realize it. 
For requests whose future layers remain offloaded, the manager writes each freshly generated KV entry directly to its designated slot in host memory, bypassing GPU capacity altogether. 
Finally, because a new placement can change the split between KV blocks on GPU and the \emph{prefetch buffer}, the manager reallocates GPU memory so that the buffer always spans a \emph{contiguous} range of blocks big enough to hold the worst-case layer transfer dictated by \autoref{eq:capacity}.
\subsection{Solver-Based Placement Planner}
\label{sec:planner}
We first formulate KV offloading as a constrained optimization problem and optimize it with an off-the-shelf solver~\cite{gurobi}. Our formulation is guided by three considerations: 
\begin{myenumerate}
    \item \textbf{SLO-awareness.} We impose a tunable upper bound on the rate of token-level SLO violations. The solver reports \emph{infeasible} when \emph{no} placement satisfies this cap. \ours{} then triggers a fallback mechanism that trims the batch and re-optimizes.
    \item \textbf{Solver overhead.} We run the solver one decode step ahead and on a separate thread, so it adds no stalls as long as its overhead stays below the decode latency. To keep it small, we shrink the search space to \revrewrite{distance-driven placements for each} request, without exhaustive searching of all layer-placement combinations.
    \item \textbf{Adaptive call frequency.} We let the solver lengthen the interval between invocations when the SLO margin is ample, and shorten it when it is tight, so it is executed only when necessary. 
\end{myenumerate}
Under these three principles, the KV offloading problem reduces to a compact ILP problem that, for each batch of requests, returns a KV placement that satisfies GPU memory limits and SLO guarantees with negligible runtime overhead.
\subsubsection{Latency Modeling}
We solve for \emph{one} decode iteration for a batch of 
$\mathcal R$ requests that traverse $L$ Transformer layers.
The solver's immutable inputs are  
(i) request-specific KV block sizes $b_r\;(r\!\in\!\mathcal R)$,  
(ii) the peak interconnect bandwidth $B$, and  
(iii) per-layer compute times $\mathit{comp}_\ell\;(\ell\!=\!1,\dots,L)$.  
Its \emph{decision variable} is the binary placement matrix
$\mathbf x=[x_{r,\ell}]\! \in\!\{0,1\}^{|\mathcal R|\times L}$, where
$x_{r,\ell}=1$ keeps the KV of request $r$ at layer $\ell$ on GPU and
$x_{r,\ell}=0$ offloads it to CPU.

Given a placement $\mathbf x$,~\autoref{alg:ps-batch} yields the
batch latency $T_{\text{batch}}(\mathbf x)$.  
The solver tackles the following optimization problem:

\[
\min_{\mathbf x\in\{0,1\}^{|\mathcal R|\times L}}
      T_{\text{batch}}(\mathbf x),
\]

\noindent subject to the GPU capacity, decode window, and token-level SLO
constraints introduced next.
\begin{algorithm}[t]
  \caption{Batch-level decode latency given a KV placement}
  \label{alg:ps-batch}
  \begin{algorithmic}[1]
    \State \textbf{Input:} KV placement $x_{r,\ell}$, block sizes $b_r$, peak memory bandwidth $B$, per layer compute times $\mathit{comp}_\ell$
    \State \textbf{Output:} batch latency $T_{\text{batch}}$
    \State $\emph{All}      \gets \{(r,\ell) \mid x_{r,\ell}=0\}$  \Comment{every required transfer}
    \State $\emph{Finished} \gets \varnothing,\;
           \emph{Pending}  \gets \emph{All},\;
           \emph{Cur}      \gets \varnothing$
    \State $\mathit{Stall}_{\text{tot}}\gets 0,\;
           \mathit{Comp}_{\text{tot}}\gets 0,\;
           \mathit{comp}_{\text{prev}}\gets 0$
    \For{$\ell = 1$ \textbf{to} $L$} \Comment{boundary \emph{before} layer $\ell$}
        \State $Q \gets \textsc{CanStart}(\emph{Pending},\,\emph{Cur})$
        \State \textsc{Promote}$(Q,\text{from}=\emph{Pending},\text{to}=\emph{Cur})$
        \State $\emph{DestL} \gets
               \{\,e\in\emph{Cur} \mid e.\text{layer}=\ell\,\}$  \Comment{tasks blocking layer $\ell$}
        
        \State sort \emph{Cur} ascending by \texttt{size}
        \State $\mathit{elapsed}\gets 0$;\;
               $n \gets |\emph{Cur}|$;\;
               $\mathit{stall}_\ell\gets 0$;\;
               $i\gets1$
        
        \While{$\emph{DestL} \neq \varnothing$}
            \State $\beta \gets B/n$
            \State $\mathit{need} \gets \emph{Cur}_i.\text{size}/\beta$
            \State $\mathit{elapsed} \gets \mathit{elapsed} + \mathit{need}$
            \If{$\mathit{elapsed} > \mathit{comp}_{\text{prev}}$}
                \State $\mathit{stall}_\ell \gets
                       \mathit{elapsed} - \mathit{comp}_{\text{prev}}$
            \EndIf
            \If{$\emph{Cur}_i \in \emph{DestL}$}
                \State $\emph{DestL} \gets
                       \emph{DestL} \setminus \{\emph{Cur}_i\}$
            \EndIf
            \State remove $\emph{Cur}_i$ from \emph{Cur};\;
                   $n \gets n - 1$;\;
                   $i \gets i + 1$
        \EndWhile
        \State $\mathit{Stall}_{\text{tot}} \gets 
               \mathit{Stall}_{\text{tot}} + \mathit{stall}_\ell$
        \State $\mathit{Comp}_{\text{tot}}  \gets 
               \mathit{Comp}_{\text{tot}}  + \mathit{comp}_\ell$
        \State $\mathit{comp}_{\text{prev}} \gets \mathit{comp}_\ell$
    \EndFor
    \State $T_{\text{batch}} \gets \mathit{Comp}_{\text{tot}} + \mathit{Stall}_{\text{tot}}$
  \end{algorithmic}
\end{algorithm}
\paragraph{GPU Capacity Bound.}
The sum of all blocks resident on GPU is bounded by the total GPU budget, i.e., the total number of KV blocks that the GPU can hold at any time. We denote this budget by $B_{\mathrm{GPU}}$, which consists of all blocks we keep on GPU and a dedicated \emph{prefetch buffer} reserved for blocks that are fetched just-in-time for the next layer. Formally,
\begin{equation}
   \sum_{r\in\mathcal{R}}\sum_{\ell=1}^{L}
        b_{r}\,x_{r,\ell}
   \;+\; B_{\text{buf}}
   \;\le\; B_{\mathrm{GPU}},
   \label{eq:capacity}
\end{equation}
where the prefetch buffer must be large enough to hold all offloaded blocks required
by any \emph{one} layer:
\[
   B_{\text{buf}} \;=\;
   \max_{\ell\in\mathcal{L}}
      \sum_{r\in\mathcal{R}}
         (1 - x_{r,\ell})\,b_{r}.
\]
\paragraph{Bounded SLO Violations.}

We formulate the SLO as a hard feasibility check. Concretely, we require that the number of SLO violations of a decode step across a batch of requests $\mathcal R$ be at most a tunable cap $\alpha$:
\[
   \sum_{r\in\mathcal R}\text{sloFail}_r \;\le\; \alpha,
\]
where $\text{sloFail}_r$ denotes whether a request $r$ violates the SLO in the current decode step and $\alpha$ controls how many violations the solver can tolerate across the batch $\mathcal R$. Under this constraint, latency becomes a meaningful objective: many placements satisfy the same SLO bound but incur markedly different latencies, with the lower one being more preferable for exploiting Token-Deposit.
If we set $\alpha$ to 0, a plan that leads to any SLO violation across requests will be rejected.
If no placement satisfies 
the constraint, the model is declared \emph{infeasible}; the runtime then pauses a request and re-optimizes (see \autoref{sec:pause_resume}).
Because pausing a request causes at most one violation in each decode step across the batch, we count each pause as one potential violation and set $\alpha$ to 1 in \ours.
Proactive pausing is
preferable in this scenario as it frees both GPU memory and bandwidth, allowing the remaining requests to run under a placement that achieves strictly lower latency than any placement that tries to keep all requests alive.

\paragraph{Self-Determined Decode Window.}
Invoking the solver at \emph{every} decode step would neutralize the exact latency we try to minimize. Instead, we let the solver itself decide how long it can safely run unchecked by introducing an integer variable $\Delta\in[\Delta_{\min},\,\Delta_{\max}]$, which we denote as the \emph{decode window}.
Formally, the window length interacts directly with the SLO feasibility constraint. Instead of enforcing the token-level SLO constraint for the current step, we enforce it on the average over the next $\Delta$ steps:
\[
\frac{1}{\Delta}
\sum_{t}^{\,t+\Delta-1}\;
\sum_{r\in\mathcal R}\text{sloFail}_r
\;\le\;  \alpha.
\]
Mechanically, if the SLO is tight, the solver chooses a small $\Delta$ so that the average is dominated by the imminent steps, the KV placement will be re-optimized sooner. On the other hand, if the latency margin is ample, the solver chooses a larger $\Delta$ and amortizes minor violations over a longer period, avoiding redundant re-optimizations. The lower bound $\Delta_{\min}$ is a tunable parameter which guarantees that the solver is never invoked more frequently than every $\Delta_{\min}$ iterations, as a way to keep the solver time bounded.

\paragraph{Scheduling KV Prefetching.}
To assess whether a batch meets its token-level SLO, we first need an
accurate estimate of compute time and the data-transfer stalls
introduced by a given placement. A placement
matrix $\mathbf{x}$ specifies, for every request $r\in\mathcal R$ and
layer $\ell\in\{1,\dots,L\}$, whether that layer's KV blocks
reside on GPU ($x_{r,\ell}=1$) or must be fetched from CPU
($x_{r,\ell}=0$). The aggregate blocks that must cross the
PCIe bus in one decode iteration are
\[
\emph{blocks\_to\_fetch}=
   \sum_{r\in\mathcal R}\sum_{\ell=1}^{L} b_r\,(1-x_{r,\ell}).
\]
Because LLM layers execute strictly sequentially, a layer can begin only when both of its inputs are ready: 1) the outputs produced by its previous layer, and 2) the KV cache for that layer. A \emph{stall} occurs when the KV blocks are still in flight while the previous layer has finished its computation. The compute units must idle until the fetching of those blocks completes. Because layers are serialized, a stall at any layer pushes back the start time of all subsequent layers, propagating delay throughout the rest of the decode iteration. 
To best overlap data transfers with computation, we launch prefetches greedily as soon as their preconditions are satisfied. Because a request can have at most one layer's KV in flight at any time, it needs only a single asynchronous data transfer stream. For any layer, the total number of active prefetch streams is bounded by the number of requests $|\mathcal{R}|$. A request's stream is \emph{active} only if all of the following hold: \textbf{(1) Current layer's KV is on GPU}, i.e. $x_{r,\ell}=1$. Otherwise, the execution of layer $\ell$ must reuse the request's prefetch buffer and launching another transfer could overwrite data in use. \textbf{(2) Some later layer's KV is offloaded}. There exists an \(\ell' > \ell\) with \(x_{r,\ell'}=0\); otherwise, all required transfers for this decode iteration are already complete.
\textbf{(3) The transfer is still in flight}. The stream that fetches the KV cache for layer~\(\ell'\) has not finished when layer~\(\ell\) begins execution.

We assume all active prefetch streams divide the available bandwidth evenly. When the smallest transfer finishes, its share is redistributed equally among the remaining streams.~\autoref{alg:ps-batch} applies this greedy bandwidth-sharing rule to compute the batch-level decode latency for a given KV placement. 
The batch-level decode latency $T_{\text{batch}}$ is the optimization objective of the Planner. 
\subsubsection{Pruning the Search Space}
A completely free layout would force the solver to explore
\(2^{|\mathcal R|\,L}\) combinations, one binary choice for every request-layer pair. 
An exhaustive search is prohibitively expensive, 
whether it is performed offline or online. Therefore, a practical system requires a principled way to prune the search space and focus only on candidates that matter.
We achieve this in two steps.
First, we drop the uniform requirement. As discussed
in \autoref{sec:prev_methods}, enforcing a uniform offload strategy for the entire batch lets long requests monopolize GPU memory and push shorter or newer sequences into SLO violations. Second, we retain the distance-driven policy itself. 
Grouping layers this way aligns identical compute times with identical KV transfers: offloading every \(N\)-th layer turns execution into repeating \(N\)-layer groups whose compute, communication, and stall patterns are identical, yielding a steady, predictable runtime.
The per-request alternative to the $2^{L}$ search is:
\[
D(L)=
\Bigl|
  \bigl\{\,
       \lfloor L/k\rfloor \;\bigm|\; k=2,\dots,L
  \bigr\}
\Bigr|
\;\le\;
2\lfloor\sqrt{L}\rfloor-1,
\]
and the search space size drops from \(2^{|\mathcal R|\,L}\) to
\(D(L)^{|\mathcal R|}\). As an example, with \(L=32\) and \(|\mathcal R|=4\), we have \(D(32)=9\) possible KV placements. Dropping the distance
constraint altogether would raise the space to
\(2^{128}\!\approx 3.4\times10^{38}\) placements, whereas keeping it but
letting each of the \(|\mathcal R|=4\) requests choose its own distance
yields only \(9^{4}=6\,561\) possibilities (30 orders of magnitude smaller).
\paragraph{Solver Overhead.} The ILP nature of the solver makes it NP-hard, which means the worst-case execution time grows exponentially. \ours{} hides this overhead in practice with three measures: 1) The search space is kept small by considering only distance-driven placements for each request, without exhaustive searching of all layer-placement combinations.
\revmajor{2) The solver is launched on the CPU ahead of time to prevent GPU computation delays. We determine when to launch it by comparing the solver overhead to the time per decoding step. With the pruned search space, the solver overhead is usually less than one decoding step. Therefore, we trigger the solver one step ahead and adjust each request’s token count by adding one token to align with the target step.}
3) The solver is invoked sparingly, based on the solver-determined window or by the \emph{Adaptive Controller} when batch conditions change.

\begin{figure*}[t]
     \centering
     \begin{minipage}[t]{0.79\linewidth}
         \centering
         \includegraphics[width=\linewidth]{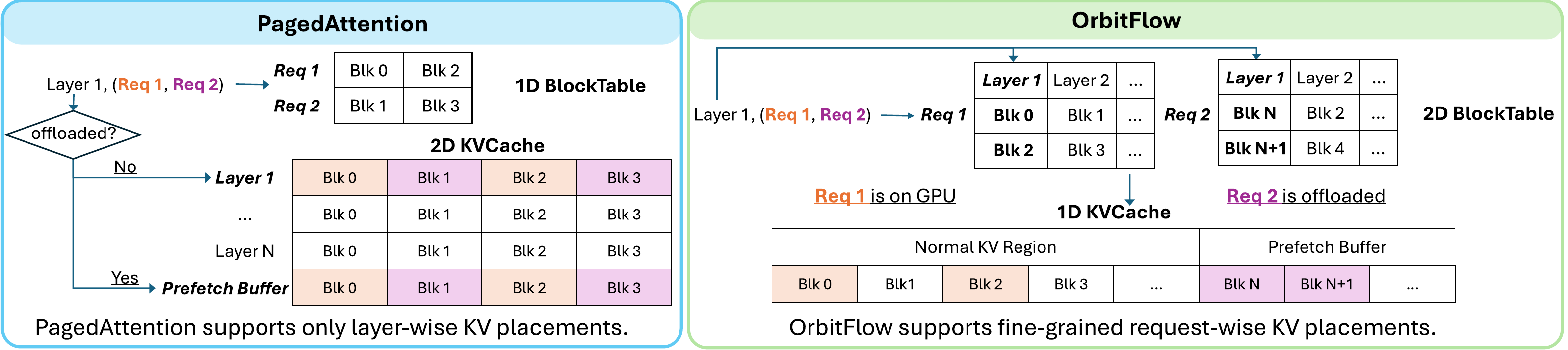}
        \caption{Illustration of flattened 1D KV cache layout in \ours{} modified from PagedAttention.}
        \Description{}
        \label{fig:1dcache}
     \end{minipage}
     \hfill
     \begin{minipage}[t]{0.2\linewidth}
         \centering
         \includegraphics[width=\linewidth]{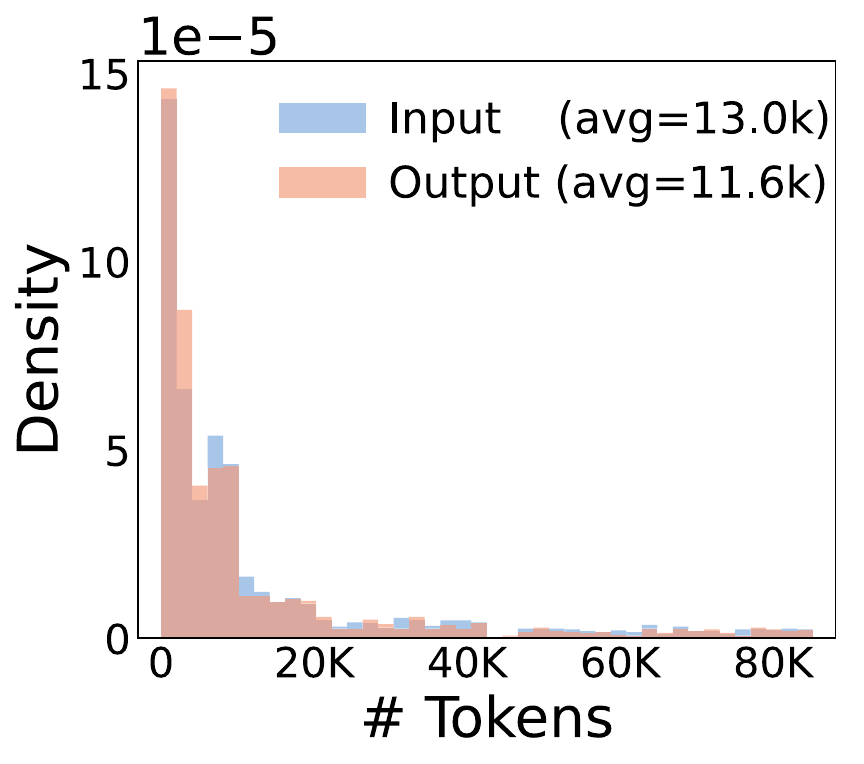}
	\caption{The length distribution of traces.}
    \Description{}
	\label{fig:workload_dist}
     \end{minipage}
    \vspace{-0.2em}

\end{figure*}


\subsection{Adaptive Controller}
\label{sec:controller}
The \emph{Adaptive Controller} monitors the execution signals, re-invokes the \emph{Planner} when necessary, and provides a fallback mechanism when the Planner deems the current batch infeasible. 
\paragraph{Solver Invocations.}
\label{sec:dyn_reconf}
The solver-chosen decode window \(\Delta\) already guards against
token-dimension drift: it guarantees that the current KV
placement remains optimal for at least the next \(\Delta\) decode
iterations \emph{provided} (i) the requests in the current batch stay unchanged and
(ii) the profiled performance model is accurate. We keep an online monitor that captures signals to re-invoke the solver whenever either assumption breaks earlier than expected:
\begin{myenumerate}
\item {Profile-mismatch trigger.};
After each decode iteration, \ours{} compares the iteration latency predicted by~\autoref{alg:ps-batch} with the latency it actually observes. If the absolute difference exceeds a configurable fraction of the predicted value, \ours{} interprets the gap as evidence of unexpected GPU or bandwidth contention. It then updates the profiled per-layer compute times and effective bandwidth using an exponentially weighted moving average and immediately re-runs the solver with the refreshed profile.
\item {Batch-change trigger.}\;
The arrival of a new request or completion of an existing one alters
the batch composition~\(\mathcal R\), causing a
batch-dimension mismatch between the current placement
and the new batch composition. The solver is re-run with
the updated batch so that KV placement reflects the fresh load mix.
\end{myenumerate}

Both triggers are infrequent. Profile deviation surfaces only under contention, and batch changes can occur at most once per request completion or arrival.

\paragraph{Pause-Resume Fallback.}
\label{sec:pause_resume}
When the solver reports \emph{infeasible} under the
GPU capacity and SLO bounds, \ours{} abandons the ``serve-all'' stance and proactively \emph{pauses} one request instead of forcing an extreme offload pattern that would cause a cascade of SLO violations. The victim is the request with the largest sum of KV blocks and deposited tokens. Selecting such a request prevents any single long request from monopolizing GPU memory while masking as many SLO violations as possible for the victim. Unlike reactive \emph{preemption}, which is triggered when the GPU memory constraint is violated and therefore drops a request's entire KV cache in one shot, \ours{} pauses a request \emph{before} memory becomes tight.
Because the GPU still has headroom at the moment of pausing, \ours{} evicts the request's KV blocks \emph{on demand and at the layer-level}: its KV blocks on GPU are dropped layer by layer when their space is actually needed. This on-demand eviction makes a later resume much cheaper than repatriating an all-or-nothing dump, requiring at most those evicted blocks.
\revmajor{A request can be paused and resumed multiple times. \ours{} resumes a paused request only after some in‑flight requests have completed and a feasible plan exists (i.e., the transient overload has cleared).}
\revmajor{If the server becomes overloaded again, \ours{} chooses a new or previously paused request as a victim. This gating mechanism ensures no single request monopolizes resources and starves others.}

\revmajor{\subsection{Extension to Distributed Inference}
\label{sec:4_5}

Distributed inference becomes necessary when KV offloading on a single GPU cannot fully relieve memory pressure. We currently focus on single-node, multi-GPU serving where a single model is configured to use all GPUs within a node. In this scenario, tensor parallelism (TP)~\cite{megatron-ml}, which symmetrically shards weights and computations across the GPUs, is the most suitable among common parallelism strategies. Pipeline parallelism (PP)~\cite{gpipe} can also be adopted to reduce per-GPU memory demand, but it depends on microbatches to hide pipeline bubbles. Unlike training, serving handles requests that arrive stochastically, so microbatches rarely stay full and pipeline stages can sit idle. 

We extend \ours{} to support TP and validate its effectiveness experimentally in~\autoref{fig:tp_tbt}. With symmetric sharding and shared memory bandwidth, a KV placement computed for one worker is expected to be optimal for all workers. \ours{} runs the solver on one worker and broadcasts the resulting placement to the others. The Runtime Executor of each worker installs the placement synchronously, thereby maintaining consistent KV placement across workers. 
We note that PP and data parallelism (DP)~\cite{DP} are typically more desirable choices than TP when serving scales to multiple nodes with slower interconnects. Since PP incurs lower communication costs than TP, a single model instance can be organized with PP across nodes and TP within each node. DP is used to scale out further by coordinating multiple model instances via a global scheduler. We leave extension to these scenarios as future work.
}

\section{Implementation}

\label{sec:impl}

We build \ours{} on top of vLLM v0.6.6~\cite{vllm} with three core changes. First, we enable request-wise KV placement by modifying vLLM's cache engine to flatten the original per-layer KV layout into one contiguous 1D layout, indexed by a layer-aware 2D table. As illustrated in \autoref{fig:1dcache}, this allows flexible allocation of a different set of layers for each request, unlike vLLM's default KV cache management, which uniformly moves every $N^{\text{th}}$ layer for all requests. We also embed the \emph{KV manager} in vLLM's cache engine, which pulls the placement decision from the Planner and issues necessary data movements to realize the placement. Second, we enable \emph{Pause-Resume} by tweaking the scheduler to temporarily drop a running request. To optimize Pause-Resume, we implement the \emph{removable cache} to manage paused requests efficiently. Rather than immediately offloading all GPU-resident layers of paused requests, we incrementally move layers to CPU only when additional GPU space is needed, reducing data transfer overhead and resume latency. Third, the \emph{Placement Planner} runs in a separate thread and always solves one decode iteration ahead. The cache engine pulls the finished plan in the next iteration and applies the changes, without device-wide synchronization. 


\section{Evaluation}
\label{sec:evaluation}
We evaluate \ours{} against baseline offloading systems, including DeepSpeed~\cite{deepspeed}, FlexGen~\cite{flexgen} and SLO-aware Offloading~\cite{selectn}, on ShareGPT-derived synthetic traces across different arrival rates. Our evaluation shows that \ours{} outperforms the baselines, achieving 62\% and 66\% higher TPOT and TBT SLO attainment, and up to 3.3$\times$ higher throughput. Additionally, \ours{} consistently provides performance improvements under various context lengths, batch sizes, and distributed serving scenarios.

\subsection{Experimental Setup}
\paragraph{Models and Hardware Settings.}

We evaluate \ours{} using LLaMA3 models~\cite{llama3}. Our default configuration uses the 8B model on a single NVIDIA RTX A5000 (24GB) GPU with 384GB of host memory over PCIe 3.0$\times$16.
For longer contexts up to 128K tokens, we also report results on LLaMA3-70B using a 4-GPU node with RTX A6000 (48GB) and 256GB of host memory over PCIe 4.0$\times$16.

\paragraph{Workloads.}
\label{eval:workload}
Because publicly available production traces for long-context serving are limited, we synthesize workloads by sampling requests from ShareGPT~\cite{sharegpt}. We use Poisson arrivals with varying arrival rates
to emulate real-world LLM serving scenarios. The generated traces span a wide spectrum of short to long inputs and outputs, as illustrated in~\autoref{fig:workload_dist}. 
Unless otherwise specified, we use a
batch size of 4 and limit the total sequence length per batch to 32K tokens.\footnote{On a single \texttt{RTX~A5000} (24GB), an 8B model uses about 16GB of GPU memory, and PyTorch runtime consumes $\sim$6GB. The remaining $\sim$2GB can hold roughly 16K tokens. With offloading, the system can accommodate up to 32K tokens per batch without memory oversubscription.}
For design validations and sensitivity studies, we fix the arrival rate at $0.97\,\mathrm{req}/\mathrm{min}$ and vary individual parameters including burstiness of arrivals, batch size, and context length.


\paragraph{Key Metrics.}
We focus on two complementary latency SLOs.
TBT measures the latency between consecutive output tokens and thus captures user-perceived interactivity at the token level. 
TPOT measures the average per-token latency.
We set the base SLO by profiling the decoding latency of the longest request that can be served without offloading, and scale it to create progressively looser budgets for fair comparison across latency constraints. Unless otherwise specified, experiments use an SLO scale of 1.5, representing a mid-level constraint.
We also report throughput, requests completed per minute, and end-to-end (E2E) latency, and the average per-request completion time, to capture overall serving efficiency.

\paragraph{Baselines.}
We consider five baseline KV cache offloading strategies, all implemented on the same codebase for a fair comparison.
\begin{myenumerate}
    \item \textbf{DeepSpeed-Inference}~\cite{deepspeed-inference} offloads all KV caches to the CPU and retains only the current layer's KV cache in GPU. It overlaps data transfers with computation by prefetching the next layer's KV cache during the current layer's execution. 
    \item \textbf{FlexGen}~\cite{flexgen} maximizes 
throughput by profiling a fixed, optimal KV placement offline, assuming peak GPU performance. It does not consider SLO constraints or reconfigure at runtime.
    \item \textbf{FlexGen${+}$} is an enhanced version of FlexGen that dynamically reconfigures its KV placement when requests join and leave, allowing it to adapt to batch-dimension mismatches. 
    \item \textbf{SLO-aware Offloading}~\cite{selectn} adjusts KV cache placement based on offline profiling
    while maximizing throughput within SLO constraints.
    It still employs uniform offload distances across requests, and reconfiguration is confined to batch boundaries, similar to FlexGen${+}$. 
    \item \textbf{Dynamic Heuristic} is a variant of \ours{} that replaces the solver-based planner with a heuristic that minimizes offloaded layers to reduce communication costs. It reconfigures at the iteration level but lacks the solver's fine-grained optimization.
\end{myenumerate}
DeepSpeed-Inference and FlexGen use \emph{static} placements determined offline, while the remaining three baselines \emph{adaptively} reconfigure at runtime.

\begin{figure*}[t]
	\centering
	\includegraphics[width=0.96\textwidth]{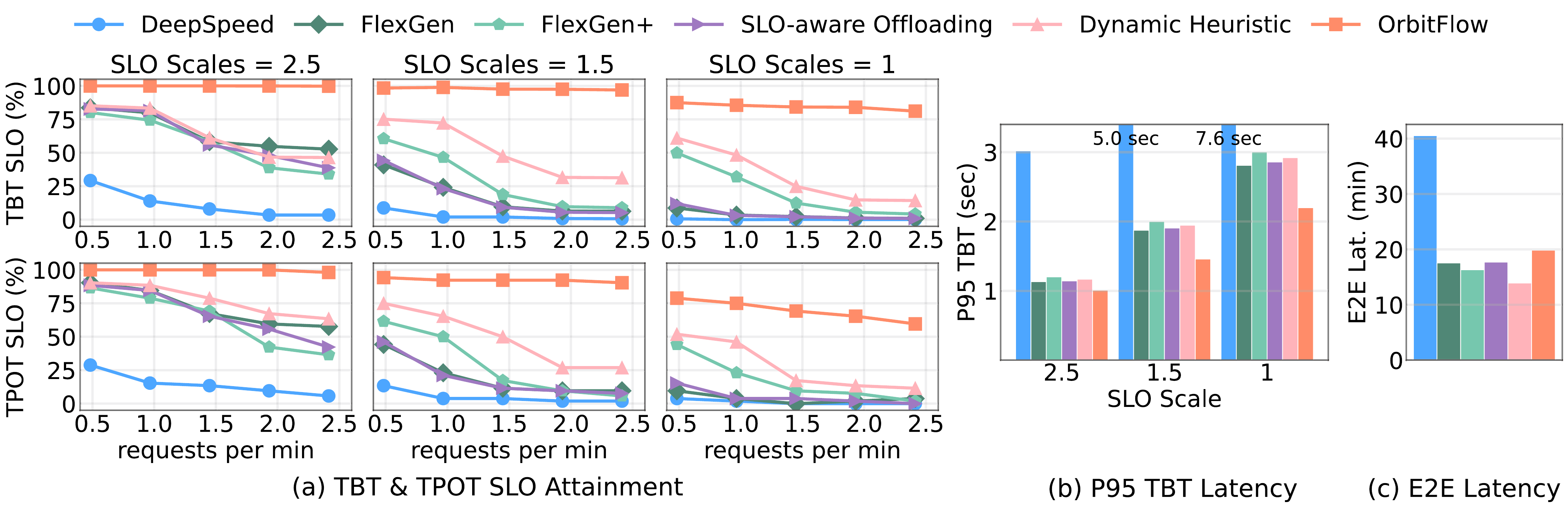}
	\vspace{-0.4em}
	\caption{(a) TBT and TPOT SLO attainment across arrival rates and SLO scales. Y-axis indicates the SLO attainment, and X-axis indicates the arrival rate. \revminor{(b) Comparison of P95 TBT latency across varying SLO scales.} \revmajor{(c) Comparison of end-to-end latency.}}
    \Description{}
	\label{fig:arrival_rate_tbt_tpot}
	\vspace{0.2em}
\end{figure*}


\subsection{Performance Analysis}
\label{sec:eval:perf}


\paragraph{SLO Attainment.}
We evaluate the effectiveness of \ours{} by examining its performance in terms of TBT and TPOT SLO attainment compared to the baselines under varying request arrival rates and SLO scales, as illustrated in~\autoref{fig:arrival_rate_tbt_tpot}a.

We observe that \ours{} consistently outperforms the baselines in both  TBT and TPOT SLO attainment, maintaining over 75\% TBT attainment under high request rates and a tight SLO scale. 
Among the \emph{static} methods, DeepSpeed-Inference shows the weakest performance, as its layer-wise offloading stalls computation with frequent KV transfers that cannot be hidden within a single layer's compute time.
FlexGen does not reconfigure at runtime or account for SLO
constraints, so its static placement becomes increasingly suboptimal as the KV cache grows during decoding.
The \emph{adaptive} baselines perform better but remain limited.
FlexGen${+}$ improves over FlexGen by reconfiguring whenever the batch composition changes, accounting for batch-dimension mismatch. 
SLO-aware Offloading adjusts offload distances in response to the SLO target---offloading more aggressively under loose constraints
and reducing offload distances as the SLO tightens---but still
applies uniform distances across requests, limiting its adaptability
when requests in the same batch have different KV cache sizes.
Dynamic Heuristic performs best among the baselines, yet its greedy
approach lacks the precision of solver-based planning to sustain
high attainment.

Under relaxed latency constraints (SLO scale=2.5), most baselines except DeepSpeed-Inference maintain stable attainment at lower arrival rates. However, attainment drops to around 50\% as the arrival rate increases. 
This degradation accelerates when the SLO scale is tightened further. 
\ours{} experiences the same trend, but demonstrates considerably better robustness. Its solver-driven, adaptive fine-grained approach adapts to both token-dimension drifts and batch-dimension mismatch, the sources of performance degradation that static policies cannot accommodate. 
TPOT attainment penalizes any per-token latency spike more heavily, as TPOT evaluates average latency at the request level. 





\begin{figure}[t]
	\centering
	  \vspace{2em}
	\includegraphics[width=0.475\textwidth]{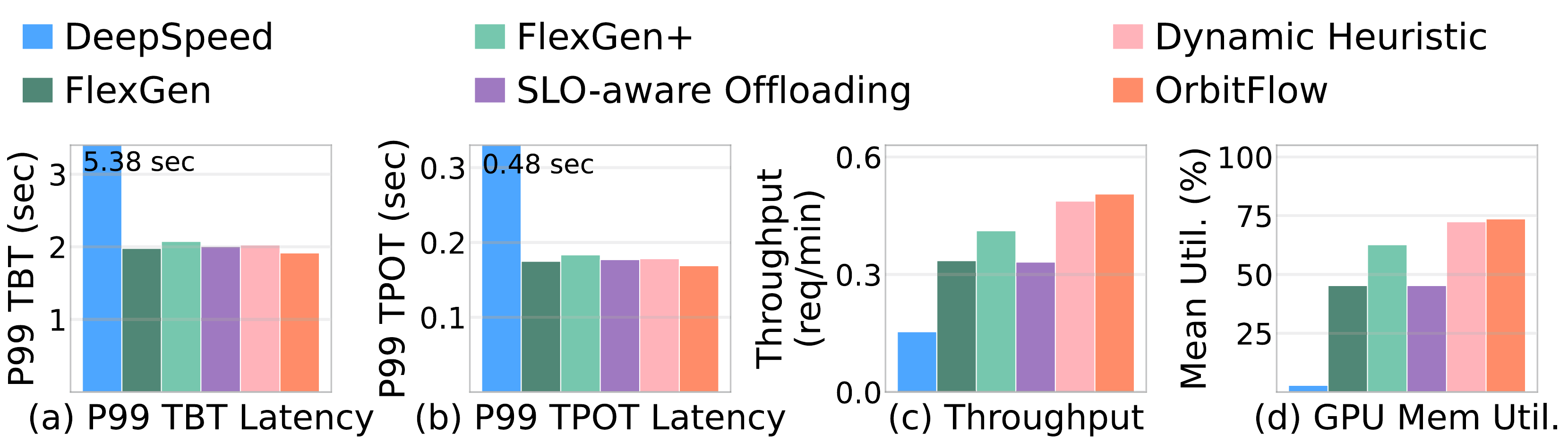}
	\caption{Comparison of (a) P99 TBT latency. (b) P99 TPOT latency. (c) Throughput. (d) Average GPU memory utilization. }
    \Description{}
	\label{fig:p99_tbt_tpot_mem}
\end{figure}

\paragraph{Tail Latency (P95 and P99).}
Beyond average SLO attainment, we examine how \ours{} performs under worst-case conditions.
\autoref{fig:arrival_rate_tbt_tpot}b reports the P95 latency
across SLO scales. \ours{} consistently demonstrates the lowest latency. At an SLO scale of 1, 
\ours{} achieves 21.7\% and 68.6\% lower P95 latency than FlexGen and DeepSpeed-Inference, respectively.
This gap narrows as the SLO loosens, but \ours{} maintains the lowest P95 latency across all scales.
At the extreme-tail latencies (P99) under SLO scale of 1.5, \ours{} achieves a 65\% lower P99 TBT and a 64\% lower P99 TPOT compared to DeepSpeed-Inference, as shown in~\autoref{fig:p99_tbt_tpot_mem}a-b.
Against other baselines, the improvement is a consistent 3--9\% in both metrics. The margin narrows at P99 because rare, burst-induced stalls affect all methods similarly. Nonetheless, \ours{} still delivers the lowest tail latency overall.



\paragraph{E2E Latency and Throughput.}
As shown in \autoref{fig:arrival_rate_tbt_tpot}c, \ours{}'s E2E latency is 12--21\% higher than the adaptive baselines, a direct consequence of the Pause-Resume mechanism, which defers long requests to keep the batch feasible under memory and SLO constraints. Dynamic Heuristic---which does not use Pause-Resume---achieves the lowest E2E latency among all methods, confirming that the additional E2E cost is from the deferred requests. However, this is a trade-off for system-level efficiency. \autoref{fig:p99_tbt_tpot_mem}d shows that \ours{} achieves the highest throughput, outperforming DeepSpeed-Inference by 3.3$\times$ and other baselines by 4--52\%. By deferring stragglers that would otherwise slow down the entire batch, Pause-Resume allows the remaining requests to decode faster, improving both SLO attainment and throughput at the cost of longer completion time for individual deferred requests.


\paragraph{Memory Usage.}
\ours{}'s throughput gains rely on effective use of limited GPU memory.
~\autoref{fig:p99_tbt_tpot_mem}c shows the average GPU memory utilization across methods. The static methods under-utilize GPU memory:
DeepSpeed-Inference keeps at most one layer on the GPU at a time, whereas FlexGen employs a static offloading decision based on the maximum expected batch size.  Among the adaptive methods: SLO-aware Offloading prioritizes CPU placement conservatively, resulting in utilization similar to FlexGen. FlexGen${+}$ shows improvement over FlexGen by recomputing the offload distances upon batch composition changes. Dynamic Heuristic closely matches the high utilization achieved by \ours{}, as it aggressively maximizes GPU memory usage.

\subsection{Sensitivity Studies}
\label{sec:sensitivity}
We study \ours{}'s sensitivity to different workloads and serving scenarios. We first evaluate longer contexts and multi-GPU setups under tensor parallelism, then vary arrival burstiness and batch size on the default single-GPU setup. Finally, we evaluate \ours{} under an alternative request scheduling policy.

\begin{figure*}[t]
     \centering
     \begin{minipage}[t]{0.34\linewidth}
         \centering
         \includegraphics[width=\linewidth]{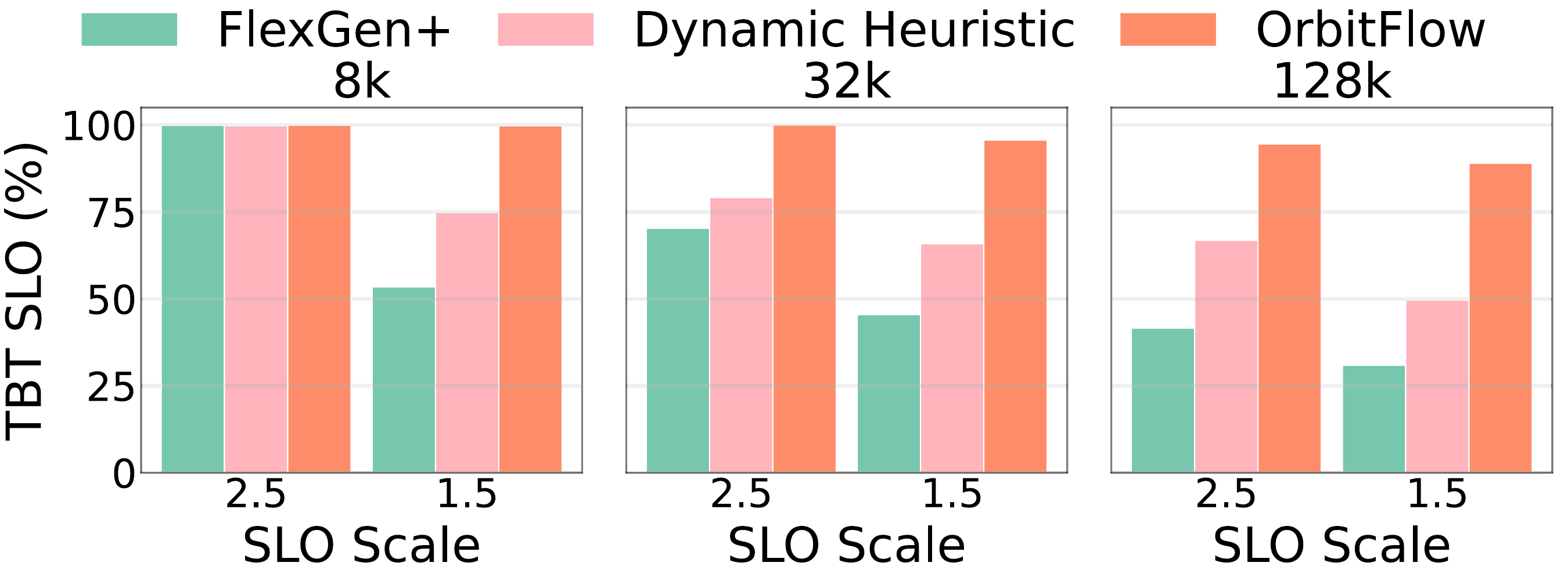}
	\caption{Comparison of TBT SLO attainment under various context lengths.}
	\label{fig:context_batch_size}
     \end{minipage}
     \hfill
     \begin{minipage}[t]{0.26\linewidth}
         \centering
         \includegraphics[width=\linewidth]{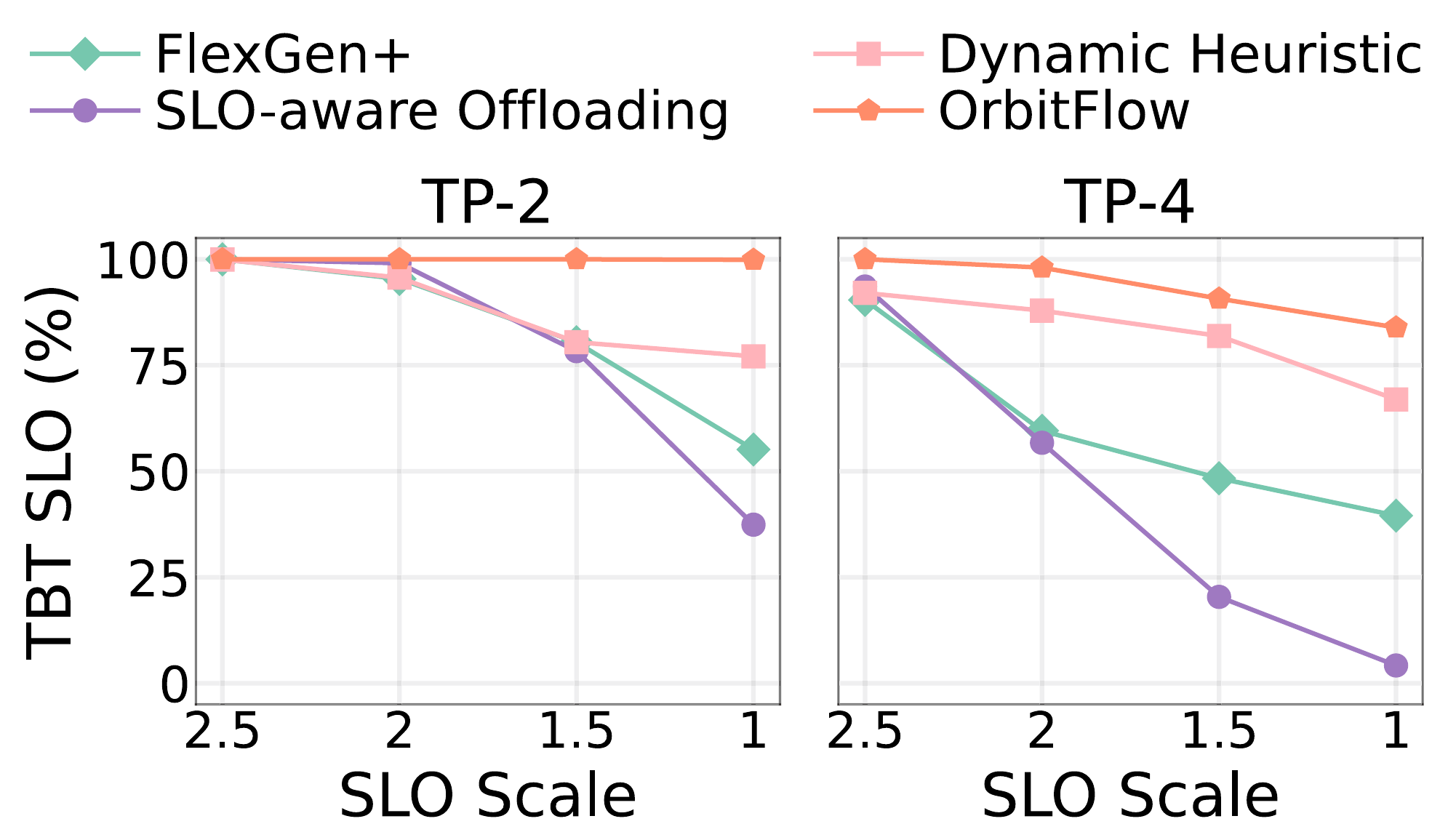}
        \caption{Comparison under 2-way and 4-way TP.}
        \label{fig:tp_tbt}
     \end{minipage}
     \hfill
     \begin{minipage}[t]{0.32\linewidth}
         \centering
         \includegraphics[width=\linewidth]{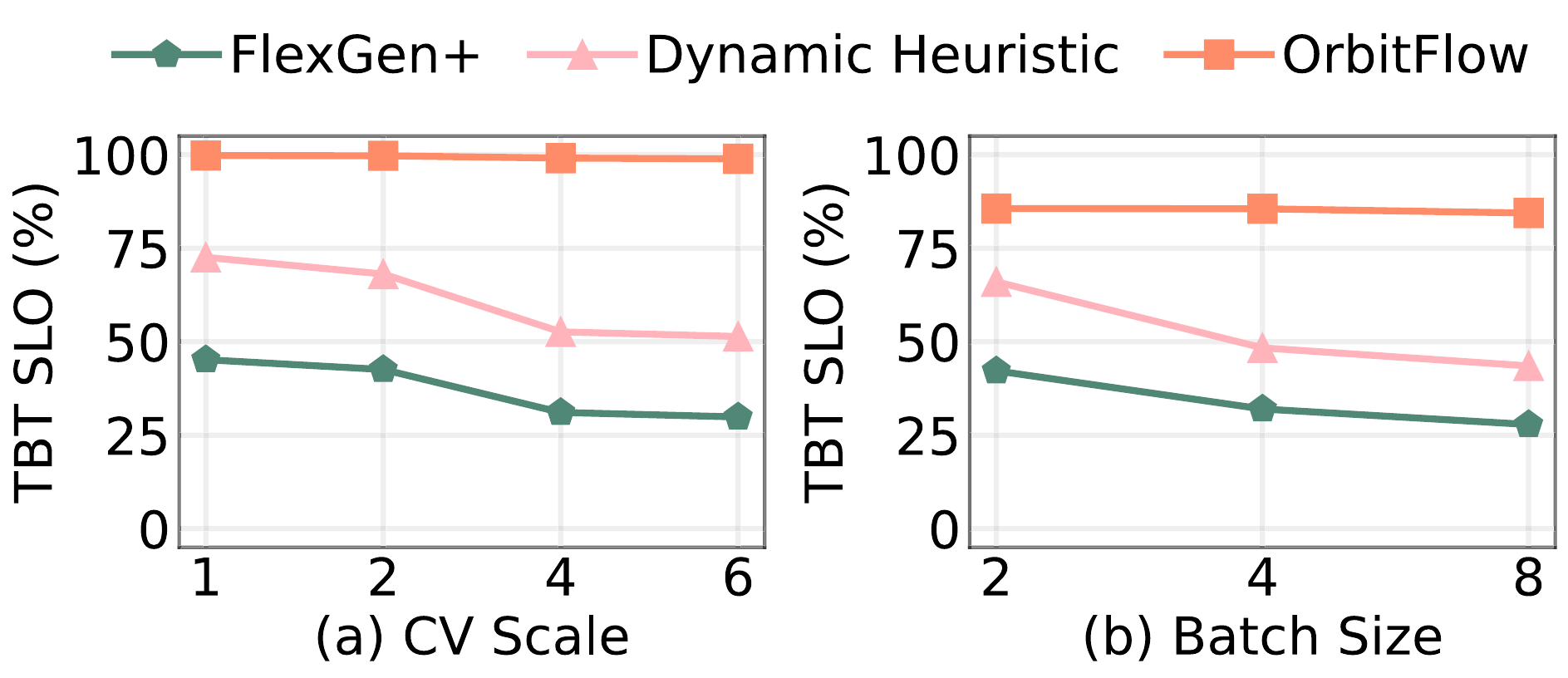}
	\caption{Sensitivity studies with varying (a) CV scales and (b) batch sizes.}
    \Description{}
	\label{fig:cv_batch}
     \end{minipage}
\end{figure*}

\paragraph{Context Length Variation.}
Increasing context length magnifies KV cache demands due to significant token-dimension drift and exacerbates batch-level imbalance, as requests with widely varying lengths are in the same batch.
\autoref{fig:context_batch_size} compares \ours{} with FlexGen${+}$ and Dynamic Heuristic---the two strongest baselines from \autoref{fig:arrival_rate_tbt_tpot}---at 8K, 32K, and 128K sequence lengths on LLaMA3-70B with 2-way TP. While FlexGen${+}$ and Dynamic Heuristic achieve near 100\% SLO attainment with 8K tokens at an SLO scale of 2.5, 
their performance degrades as context length grows, dropping to 69.2\% and 32.9\% at 128K tokens. In contrast, \ours{} consistently achieves above 85\% SLO attainment across all context lengths.
\paragraph{Tensor Parallelism.}
Tensor parallelism (TP) distributes model weights and KV caches across GPUs, requiring frequent all-reduce operations. Without NVLink, these operations rely on PCIe, directly competing with KV cache offloading and causing PCIe contention.
\autoref{fig:tp_tbt} evaluates \ours{} against FlexGen${+}$, SLO-aware Offloading, and Dynamic Heuristic under 2-way and 4-way TP. Baseline methods degrade sharply under this contention due to their static placement strategies, whereas \ours{} sustains higher TBT SLO attainment even under stringent SLOs by dynamically adapting its placement to the reduced available bandwidth.

\paragraph{Burstiness Variation.}
As shown in \autoref{fig:cv_batch}a, we vary the coefficient of variation (CV) of request arrivals to control burstiness. Higher CV produces more bursty traffic, increasing queuing delay and the risk of SLO violations. \ours{} maintains TBT SLO attainment above 90\% as CV increases, while FlexGen${+}$ and Dynamic Heuristic degrade steadily. When CV exceeds 6, all methods flatten as  bursts become intense enough to saturate the GPU. Extra burstiness only lengthens the time requests wait in the queue.

\paragraph{Batch Size Variation.}
We evaluate \ours{} under different batch sizes, as shown in \autoref{fig:cv_batch}b. Dynamic Heuristic and FlexGen${+}$ fail to scale since larger batches amplify drifts in both dimensions.
The token dimension grows more rapidly as the total KV cache size 
increases with 
batch size, and the batch dimension becomes more volatile due to frequent request completions and 
arrivals. In contrast, \ours{} effectively manages such fluctuations, maintaining consistently high performance regardless of the batch size.
\begin{figure}[t]
	\centering
	\includegraphics[width=0.47\textwidth]{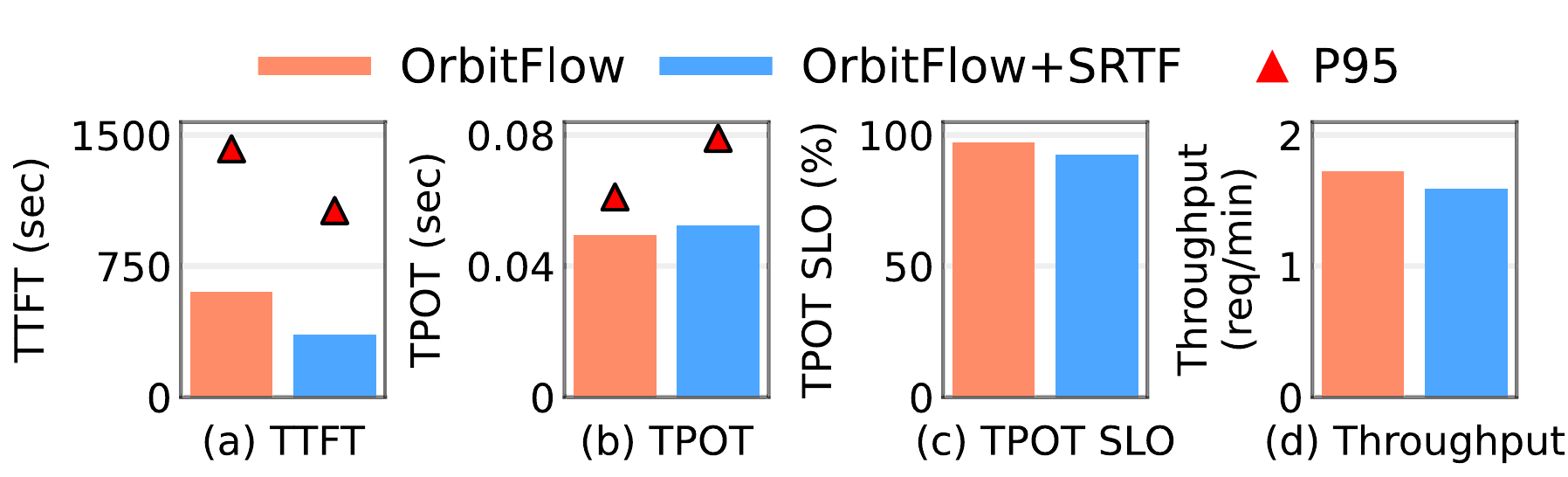}
	\caption{\revmajor{Effect of SRTF scheduling on \ours{}.}}
    \Description{}
	\label{fig:srtf}
\end{figure}
\paragraph{Effect of Request Scheduling.}\ours{} uses the first-come, first-served (FCFS) scheduling policy in vLLM by default. We further examine the performance of \ours{} under an alternative shortest-remaining-time-first (SRTF) policy, which sorts the requests in the waiting queue by their predicted output length at each admission, following~\cite{ltr}\footnote{We use the same trace as~\cite{ltr}, scaled by 6$\times$ to align with our long-context setting.}.
SRTF prioritizes requests with shorter outputs and tends to batch requests with similar output lengths, thereby helping alleviate the batch-dimension mismatch. 
This reduces the average and P95 TTFT of \ours{} by 40\% and 24\%, respectively, compared to FCFS, as shown in~\autoref{fig:srtf}a. However, SRTF inevitably clusters long requests together, forcing more KV caches to be offloaded and increasing data transfer costs, which eventually translate into long decode stalls. As a result, \ours{}+SRTF achieves TPOT SLO attainment of 92.4\%, which is 4.8\% lower than \ours{}. SRTF also leads to a 29\% increase in P95 TPOT and an 8\% drop in overall throughput. Despite these trade-offs, \ours{} maintains over 90\% TPOT SLO attainment, confirming that its runtime adaptation remains effective across different scheduling policies.

\subsection{Design Validation and Overheads}
\label{sec:design_overhead}
\begin{figure}[t]
	\centering
	\includegraphics[width=0.47\textwidth]{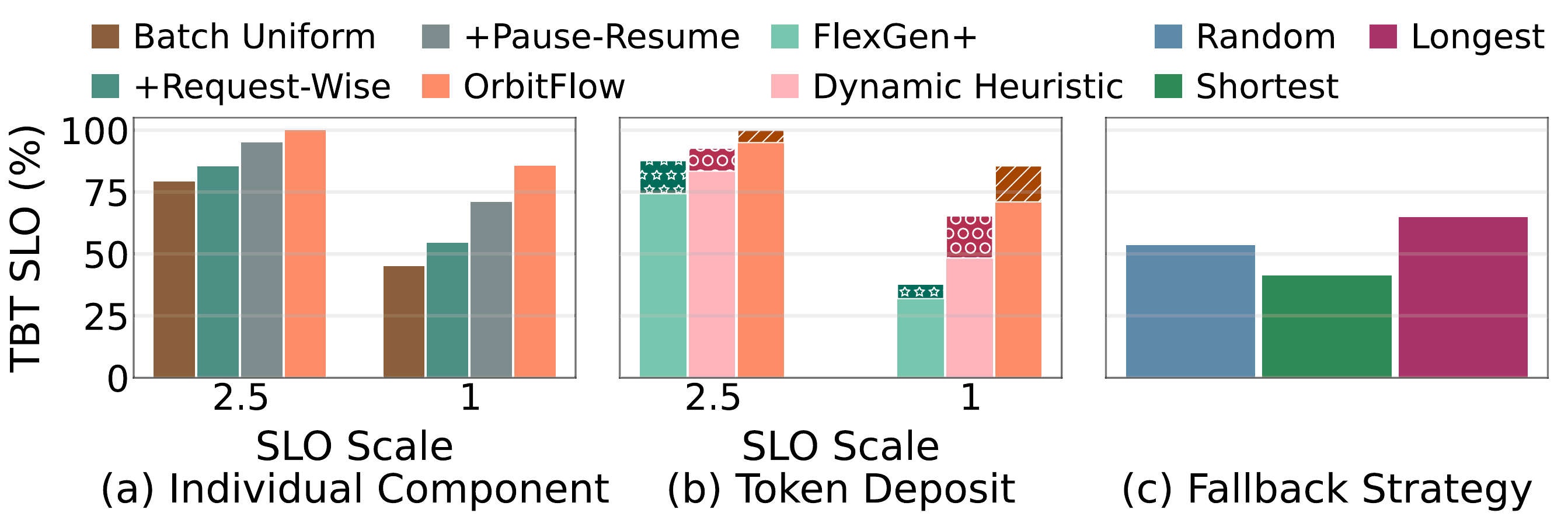}
	\caption{\revminor{(a) Impact of \ours{}'s components.} (b) Effectiveness of Token-Deposit mechanism (c) Comparison of fallback victim selection.}
    \Description{}
	\label{fig:design}
    \vspace{-0.2em}
\end{figure}
\paragraph{Impact of Individual Techniques.}
To assess the contributions of \ours{}'s core components, 
we evaluate \ours{} by adding each component incrementally, as shown in \autoref{fig:design}a.
\emph{Batch-Uniform} applies a uniform offload distance to all requests within a batch, achieving 45.1\% SLO attainment at the tightest SLO scale of 1.
Building on this, \emph{+Request-Wise} introduces fine-grained, per-request offloading decisions optimized by the solver, improving SLO attainment to 54.5\% by better balancing KV cache placement across heterogeneous requests.
Adding {Pause-Resume} further boosts SLO attainment to 71.0\% by temporarily deferring requests with large KV caches to free GPU memory for others.
Finally, \ours{} adds {Token-Deposit},
further raising SLO attainment to 85.6\%.
The same trend persists at a looser SLO scale of 2.5, with each
component contributing consistent improvements.


\paragraph{Token-Deposit on Baselines.}
\autoref{fig:design}b adds Token-Deposit to FlexGen${+}$ and Dynamic Heuristic to evaluate its generality. 
At a loose SLO scale of 2.5, Token-Deposit improves TBT attainment by 18\%, 11\%, and 5\% for FlexGen${+}$, Dynamic Heuristic, and \ours{},  respectively. As the SLO tightens, the benefit grows for Dynamic Heuristic and \ours{}, which gain a further 35\% and 20\%, as their adaptive placement generates early tokens fast enough to accumulate a buffer, whereas FlexGen${+}$ gains little benefit.

\paragraph{Fallback Strategy.}
When Pause-Resume is triggered, \ours{} must select which request to
pause. We evaluate three approaches: pausing the longest, shortest, or a randomly chosen request from the batch. We study these policies using a stress-test trace with an increased fraction of long requests to increase overload frequency and trigger more fallback events. Because KV sizes grow linearly with sequence length, pausing the longest request releases the most GPU memory, allowing the remaining requests to continue without stalls, as confirmed in \autoref{fig:design}c.

\begin{figure}[t]
	\centering
	\includegraphics[width=0.47\textwidth]{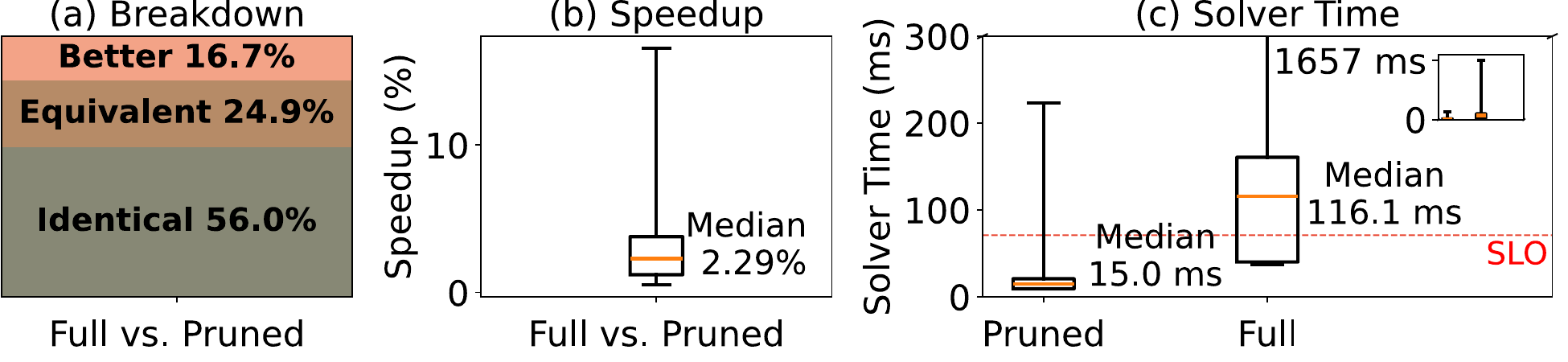}
	\caption{Effect of search space pruning of \ours{}. 
    }
    \Description{}
	\label{fig:search_space}    
\end{figure}

\paragraph{Component-wise Runtime Overhead.}
We break down the total execution time into \emph{scheduler}
(5.28\%), \emph{prefill} (7.04\%), \emph{decode} (87.23\%), \emph{KV
Manager} (0.45\%), and \emph{Placement Planner} (0.02\%), with the last
two representing overheads introduced by \ours{}. 
\autoref{tab:solver_summary_aggregated} further details the Planner's
overhead across three workloads with increasing dynamism:
\emph{Both Static} uses static batching with short outputs, limiting
changes in both dimensions;
\emph{Token Dynamic} uses static batching but with long outputs, so KV caches grow along the token dimension while the batch stays fixed;
\emph{Both Dynamic} combines long outputs with continuous batching,
varying both dimensions simultaneously.
We report the solver's accumulated share of total execution time
(E2E\%) and its per-invocation wall-time relative to one decode step
(TBT\%).
For \emph{Both Dynamic}, the solver is triggered approximately
$4\times$ and $39\times$ more often than the static workloads,
reflecting frequent changes in batch composition and KV cache size.
Even in this volatile trace, E2E\% is only 0.64\%.
Per invocation, TBT\% stays well below 100\%. As we launch the solver
one iteration ahead on a separate thread, this cost can be fully hidden under computation.
In rare cases, however, the solver exceeds one decode step, visible as the long tail in
\autoref{fig:search_space}c.

\paragraph{Search Space Pruning.}
The low overhead is enabled by pruning the search space. Our solver
restricts placements to evenly spaced offloaded layers, parameterized
by a single offload distance. For a 32-layer model, distance 4 offloads
every 4th layer (8 total) and distance 5 offloads every 5th layer
(6 total). Counts between these, such as 7 offloaded layers, require
uneven spacing and are never explored.
To evaluate the impact of this restriction, we compare our solver, \emph{Pruned}, against \emph{Full} which allows arbitrary KV placements. We sample 2000 solver inputs from the evaluation traces and compare the placements. 
\autoref{fig:search_space}a shows that 80.9\% of \emph{Full}'s
placements match \emph{Pruned} in decode latency, with 56\% identical
and 24.9\% differing only in which layers are offloaded. Only 16.7\%
are strictly better. Moreover, 98\% of the strictly better cases arise from intermediate
layer counts that even spacing skips, as we expected. The median speedup in these cases is just 2.29\% (\autoref{fig:search_space}b), while closing this gap would require $7.74\times$ more solver time (\autoref{fig:search_space}c) on average, which
would itself violate the SLO.

\section{Related Work}
\label{sec:relwork}
\begin{table}[t!]
  \centering
  \resizebox{0.42\textwidth}{!}{
    \begin{tabular}{lrrr}
      \toprule
      \textbf{Trace} & \textbf{Calls/Req.} & \textbf{E2E (\%)} & \textbf{TBT (\%)} \\
      \midrule
      Both Static    & 1.25  & 0.27 & 23.47 \\
      Token Dynamic  & 11.92 & 0.37 & 31.65 \\
      Both Dynamic   & 48.33 & 0.64 & 28.49 \\
      \bottomrule
    \end{tabular}
  }
  \vspace{10pt}
  \caption{Solver overhead across different workloads.}
  \label{tab:solver_summary_aggregated}
\vspace{-1em}
\end{table}


\revrewrite{\paragraph{KV Offloading.}
InfiniGen~\cite{infinigen} speculatively prefetches critical KV entries to minimize CPU-to-GPU transfer overhead. Neo~\cite{neo} and FlexInfer~\cite{flexinfer} extend memory capacity by offloading KV caches to host memory or SSD. These methods primarily optimize the spatial placement and movement of KV data to reduce transfer latency. \ours{} follows a similar motivation but performs placement decisions online, adapting KV allocations at runtime according to batch composition and token-level latency feedback.

\paragraph{Pruning and Compression.}
Structural pruning and quantization reduce KV cache footprints on GPUs.
StreamingLLM~\cite{streamllm} and H2O~\cite{h2o} prune less critical KV entries,
KIVI~\cite{kivi} employs asymmetric quantization, and MiniCache~\cite{minicache} merges similar KV states across layers.
Although effective in reducing memory usage, these methods risk degrading model accuracy.
}
\paragraph{\revminor{Recomputation.}}
\revminor{Recomputation rematerializes KV states rather than transferring them, but prior work has focused on the prefill stage for reusing historical prompt contexts. HCache~\cite{hcache} reconstructs KV states with activations that are lightweight in MHA models. CachedAttention~\cite{cachedattention} reuses or recomputes KV states across conversation turns to avoid redundant transfers. Although such recomputation could be applied to the decode stage, its effectiveness is still limited to MHA models. \ours{} is primarily evaluated on modern GQA models, where activations are no longer lightweight. However, in MHA models, combining KV offloading with recomputation may yield synergistic effects.}
\paragraph{Request Scheduling.}
Request scheduling techniques improve GPU utilization and batching efficiency during inference.
Orca~\cite{orca} introduces continuous batching to insert new requests dynamically as others finish,
and Sarathi-Serve~\cite{sarathiserve} integrates prefill and decode using stall-free chunked batching.
Disaggregated systems such as DistServe~\cite{distserve} decouple prefill and decode across GPU clusters, while Helix~\cite{helix} and Hetegen~\cite{hetegen} explore heterogeneous GPU graphs for model- and tensor-parallel pipelines. \ours{} complements these methods by optimizing latency within active batches through fine-grained, layer-wise KV cache placement with a solver that adapts to token- and batch-level drifts.
\section{Conclusion}
\ours{} provides an effective framework for long-context LLM serving, dynamically reconfiguring KV cache placement per request using a solver to meet stringent latency SLOs. 
Evaluations on LLaMA3-8B show that \ours{} improves TPOT and TBT SLO attainment by 62\% and 66\% under the most demanding conditions, reduces P95 TBT latency by 38\% over the best competing systems, and achieves up to 3.3$\times$ higher throughput compared to state-of-the-art offloading methods. These gains come with less than 1\% runtime overhead and persist under multi-GPU settings.


\begin{acks}
\justifying
This work was supported by SAMSUNG Research, Samsung Electronics Co., Ltd., and by the Institute of Information \& Communications Technology Planning \& Evaluation (IITP) grants (No.~RS-2025-02304554, Efficient and Scalable Framework for AI Heterogeneous Cluster Systems; No.~RS-2019-II191906, Artificial Intelligence Graduate School Program (POSTECH)), the National Research Foundation of Korea (NRF) grant (No.~RS-2024-00354947), and 
the ETRI grant (No.~26ZS1100, Development of Large-Scale Parallel Computing Technology for Generative AI), funded by the Korean government (MSIT).

\end{acks}

\balance
\bibliographystyle{plain}
\bibliography{references}

\end{document}